# Understanding and Accelerating EM Algorithm's Convergence by Fair Competition Principle and Rate-Verisimilitude Function

**Chenguang Lu**

survival99@gmail.com

**Abstract:** Why can the Expectation-Maximization (EM) algorithm for mixture models converge? Why can different initial parameters cause various convergence difficulties? The *Q*-*L* synchronization theory explains that the observed data log-likelihood *L* and the complete data log-likelihood *Q* are positively correlated; we can achieve maximum *L* by maximizing *Q*. According to this theory, the Deterministic Annealing EM (DAEM) algorithm's authors make great efforts to eliminate locally maximal *Q* for avoiding *L*'s local convergence. However, this paper proves that in some cases, *Q* may and should decrease for *L* to increase; slow or local convergence exists only because of small samples and unfair competition. This paper uses marriage competition to explain different convergence difficulties and proposes the Fair Competition Principle (FCP) with an initialization map for improving initializations. It uses the rate-verisimilitude function, extended from the rate-distortion function, to explain the convergence of the EM and improved EM algorithms. This convergence proof adopts variational and iterative methods that Shannon *et al*. used for analyzing rate-distortion functions. The initialization map can vastly save both algorithms' running times for binary Gaussian mixtures. The FCP and the initialization map are useful for complicated mixtures but not sufficient; we need further studies for specific methods.

**Keywords:** EM algorithm; mixture models; clustering; convergence proof; cross-entropy; Shannon mutual information; semantic mutual information; machine learning; rate-distortion

MSC: 68T05, 49N30, 93C41, 62B10, 68P30, 94A17, 94A34

## 1. Introduction

The task of solving a mixture model is to find model parameters according to observed incomplete data that are supposed to be produced by a true mixture model. Mixture models are important because they can be used for clustering (unsupervised learning), mixture model-based classification (semi-supervised learning), and modeling or density estimation. Gaussian mixture models are often used because Gaussian functions (with normal distributions) are symmetrical, and it is easy to calculate the model parameters. Wole [1] proposed the Gaussian mixture model in the 1960s. The Expectation-Maximization (EM) algorithm is highly-efficient and straightforward for solving mixture models. It is also used for hidden Markov models. But this paper only deals with the EM algorithm for mixture models. Hereafter, the EM algorithm means the EM algorithm for mixture models. The EM algorithm was explicitly proposed by Dempster *et al.* in 1977 [2]. Because of the EM algorithm, mixture models are widely applied to many areas, such as machine learning [3,4], remote sensing [5], electrocommunication [6], process analyses [7], and image processing [8].

Although the EM algorithm has many successful applications, it often results in slow or invalid convergence, including local convergence and boundary convergence [9]. The EM algorithm is very sensitive to initial parameters so that slightly different parameters may vastly change the speed, even convergent parameters [10,11], especially when the overlap of components is severe [12]. Therefore, most improvements focus on initial parameters, such as improvements in the Deterministic Annealing EM (DAEM) algorithm proposed by Ueda and Nakano [10], the Split and Merge EM (SMEM) algorithm proposed by Ueda et al. [13], the Competitive EM (CEM) algorithm [9], the Random swap EM algorithm [14], the cross-entropy method with EM algorithm [15], and robust EM algorithms [16]. Some improvements are made to the EM algorithm itself by different objective functions, such as in the Maximization-Maximization (MM) algorithm proposed by Neal and Hinton [17], the Variational Bayesian EM (VBEM) algorithm by Beal [18], the Expectation/Conditional Maximization Either (ECME) algorithm by Liu and Robin [19], and other algorithms [20,21]. See Appendix A for the list of abbreviations. Some improvements are made by optimizing or predicting the step sizes of parameters [23, 24]. Some improved EM algorithms can reduce 30%-60% iterations that the EM algorithm needs, as reported in [17, 21, 22]. It is a new trend to use variational methods with Bayesian inference to explain and improve the EM algorithm, such as by Beal [18], Assaad et al. [24], and Liu *et al*. [25].

This study deals with both the initialization and the objective function. It also provides a new convergence theory of mixture models because the author thought that popular convergence theories and convergence proofs were not satisfactory. There are three completely different convergence proofs of the EM algorithm for mixture models:

- **The *Q*-Proof**. The complete data log-likelihood $Q$ is used as the objective function. We need to prove that we can maximize the observed data log-likelihood $L$ by maximizing $Q$. This kind of proofs [10,26-29] is questionable. Section 2.2 will discuss further.
- **The Gradient Proof.** It was proposed by Xu and Jordan [30] and developed by others [12]. This proof builds up the connection between the EM algorithm and gradient-based approaches for maximum likelihood learning of finite gaussian mixtures. This proof should be correct and meaningful, but it is a little complicated and limited to Gaussian mixtures.
- **The Variational Proof**. Beal [18] provided a variational proof of the EM algorithm before proposing the Variational EM algorithm. This proof is not well-known. It uses the variational method to prove that the E-step increases $L$ no matter if $Q$ increases or decreases. However, to answer why the EM algorithm can converge, this proof is not sufficient.

This paper carries on Beal's Variational Proof.

The DAEM algorithm's authors [10] provided a typical $Q$-Proof. The core thought of the convergence theory basing their $Q$-Proof can be found in [10]:

"It has theoretically been shown that an iterative procedure for maximizing $Q$ over $\theta$ will cause the likelihood $L$ to monotonically increase, e.g., $L(\theta^{(t+1)}) \geq L(\theta^{(t)})$. Eventually, $L(\theta^{(t)}$ converges to a local maximum. The EM algorithm is especially useful when the maximization of the $Q$-function can be more easily performed than that of $L$."

In the above statement, $t$ means the $t$-th iteration, $L=L(\theta)=[\log P(\mathbf{X}|\theta)]/N$, and $Q=[\log P(\mathbf{X}, \mathbf{Y}|\theta)]/N$, where $N$ is the sample's size, $\mathbf{X}$ is the observed data, and $\mathbf{Y}$ is the latent data.

In Wikipedia's entry "Expectation–maximization algorithm", there is a similar statement (according to [31]):

"Expectation-maximization works to improve $Q$ rather than directly improving $L$. Here it is shown that improvements to the former imply improvements to the latter."

Let us call this theory the $Q$-$L$ synchronization theory. This theory can be traced back to the articles of Dempster *et al*. [2], Wu [26], and Little *et al*. [27]. It includes two affirmations:

**Affirmation I**: The incomplete data log-likelihood $L$ increases with the complete data log-likelihood $Q$, and we can maximize $L$ by maximizing $Q$.

**Affirmation II:** $Q$ increases in the M-step and is non-decreasing in the E-step, or $Q$ is increasing after every iteration.

However, it has never been proven that $Q$ is non-decreasing in the E-step, or $Q$ is increasing after every iteration. Although Ueda and Nakano provided proof that the E-step maximizes $Q$ (see Eqs. (5) and (6) in [10]), the author of this paper thinks that this proof is incorrect (Section 2.2 explains the reason). According to the $Q$-$L$ synchronization theory, locally maximal $Q$ can cause $L$'s local convergence. However, this conclusion is also incorrect (see Figure 1).

Ueda and Nakano later change their proof in [32] and affirm that the E-step maximizes the Shannon condition entropy $H(Y|X)$ (see this paper's Eq. (10)). However, they still use $Q$ as the objective function and adopt the DAEM algorithm to avoid $Q$'s local maxima. Most researchers still believe the above two affirmations so that we hardly see such a statement that the E-step may decrease $Q$.

After analyzing the EM algorithm by semantic information methods [33-37] and programming practices, the author of this paper believes that

- the EM algorithm is better than many researchers expect because the EM algorithm can get away from a locally maximal $Q$ to achieve a globally maximal $L$ in most cases;
- local convergence exists mainly because of small samples, slow convergence speeds, and misunderstanding (see Section 3.3);
- The $Q$-$L$ synchronization theory is incorrect because the above two affirmations are wrong.

In [35], the author proposes an improved EM algorithm, which is called the Channel Matching EM (CM-EM) algorithm later [36]. Then the author only uses continuous distribution functions as sampling distributions to test the algorithm. But the convergence proof there is improper. In [36], a section simply introduces the CM-EM algorithm and the new convergence theory.

This paper aims

- to increase our understanding of why EM algorithm can converge and why different initializations result in different convergence difficulties, and
- to improve the initializations and the EM algorithm itself for faster convergence.

For the above two purposes, this paper uses marriage competition to explain why some initial means are easy or not easy to converge validly. It proposes the Fair Competition Principle (FCP) to improve mixture models' initializations. Then it provides an initialization map to explain the FCP's application. After that, this paper introduces the new convergence theory for the EM and CM-EM algorithms. The objective function is $G/R$ or $R$-$G$, where $R$ is Shannon's mutual information [38], and $G$ is the semantic mutual information.

The new convergence theory is based on the author's semantic information theory proposed twenty years ago [33-34]. The semantic information measure can be expressed by cross-entropy and mutual cross-entropy with likelihood functions and truth functions [36,37]. This theory claims that we can maximize information efficiency $G/R$ or minimizes $R$-$G$ so that $L$ and the relative entropy $H(P||P_\theta)$ decrease in every step of the EM or CM-EM algorithm. In $H(P||P_\theta)$, $P$=$P(x)$ is the sampling distribution, and $P_\theta$=$P(x|\theta)$ is the predicted distribution or the mixture model. The new convergence theory is called the Information Efficiency Maximization (IEM) theory.

The main contributions of this study are

- It revealed that the E-step might and should decrease $Q$ in some cases.
- It clarified the causes of the EM algorithm's local convergence and provided an initialization map for selecting the suitable initial means for binary Gaussian mixture models.
- It provided the new convergence theory with a strict convergence proof for both the CM-EM and EM algorithms.

This paper will use some examples with different initial parameters to explain how these examples support the FCP, the IEM theory, and the CM-EM algorithm.

## 2. Using Cross-entropies to Explain the EM Algorithm for Mixture Models

### 2.1. *Mathematical Tools*

To use information-theoretical methods better, we use sampling distributions instead of samples to train likelihood functions.

**Definition 1.** Let $U$ be an instance or sample space, and $X$ be a random variable taking value $x$ from $U$={$x_1$, $x_2$, …, $x_m$}. Let $V$ be a label space and $Y$ be a random variable taking value $y$ from $V$={$y_1$, $y_2$, …, $y_n$}. For analytical convenience, we assume that $U$ is one-dimensional.

**Definition 2.** A sample **D** consists of $N$ examples, e.g., **D**={($x(t)$; $y(t)$), $t$=1, 2, …, $N$; $x(t)\in U$; $y(t)\in V$}. A conditional sample is $\mathbf{D}_j$={($x(t)$; $y_j$), $t$=1, 2, …, $N_j$; $x(t)\in U$}, where $N_j$ is the number of examples with $y_j$. We assume that $x(t)$ is discrete, and $m$=|$U$| is big enough so that any $x(t)$ is in $U$.

**Definition 3.** Let $\theta$ be a predictive model. For each $y_j$, there is a sub-model $\theta_j$. We define $P(x|\theta_j)$=$P(x|y_j,\theta)$, a predictive distribution, and the likelihood function of $\theta_j$. Hence, the predicted distribution of $x$ is

$$P_\theta(x)=\sum_j P(y_j)P(x|\theta_j). \tag{1}$$

We use $P(x|\theta_j)$ in most cases instead of $P(x|y_j, \theta)$ in popular methods because using $P(x|\theta_j)$ can clearly show the relationship between various cross-entropies and corresponding Shannon's entropies.

**Definition 4.** Let $P(x)$ be a sampling distribution, $P(y)$ be a label distribution for mixing proportions, and $P(x|y_j)$ be a conditional sampling distribution from the statistics of $\mathbf{D}_j$.

The cross-entropy of $X$ is

$$H_\theta(X)=-\sum_i P(x_i)\log P_\theta(x_i). \tag{2}$$

The cross-entropy has two essential properties:

**Property 1**: $H_\theta(X)$ decreases as $P_\theta(x)$ approaches $P(x)$ (left matching), and $H_\theta(X)$ reaches its minimum as $P_\theta(x)$ becomes $P(x)$. Therefore, the minimum cross-entropy criterion is equivalent to the maximum likelihood criterion [39].

**Property 2**: $H_\theta(X)$ may decrease or increase as $P(x)$ approaches $P_\theta(x)$ (right matching).

Property 2 is listed because some people mistake that the right matching also decreases cross-entropy or increases the likelihood.

The relative entropy or the Kullback-Leibler (KL) divergence is

$$H(P||P_\theta)=\sum_i P(x_i)\log\frac{P(x_i)}{P_\theta(x_i)}=H_\theta(X)-H(X), \tag{3}$$

where $H(X)$ is the Shannon entropy of $X$. Akaike in 1974 [40] proved that the maximum likelihood criterion is equivalent to the minimum KL divergence criterion because

$$L(\mathbf{X}|\theta)=[\log P(\mathbf{X}|\theta)]/N= -H_\theta(X), \tag{4}$$

and $H(X)$ does not change with $\theta$.

The cross-entropy of $X$ for given $y_j$ is

$$H(X|\theta_j)=-\sum_i P(x_i|y_j)\log P(x_i|\theta_j). \tag{5}$$

Following Akaike, we can also prove that log-likelihood $L(\mathbf{X}|\theta_j)$ for given $y_j$ is equal to the negative cross-entropy $-H(X|\theta_j)$ under IID (Independent and Identically Distribution) hypothesis:

$$\begin{aligned} L(\mathbf{X}|\theta_j)&=\frac{1}{N_j}\log P(\mathbf{X}|\theta_j)=\frac{1}{N_j}\log P(x(1),x(2),...,x(N_j)|y_{\theta j}) \\ &=\frac{1}{N_j}\log\prod_i P(x_i|\theta_j)^{N_{ji}}=\sum_i P(x_i|y_j)\log P(x_i|\theta_j)= -H(X|\theta_j). \end{aligned} \tag{6}$$

It is easy to prove that when $P(x|\theta_j)$= $P(x|y_j)$, the likelihood $L(\mathbf{X}|\theta_j)$ reaches its maximum, and the cross-entropy reaches its minimum. The EM algorithm's objective function is $Q$=[log$P(\mathbf{X}, \mathbf{Y}|\theta)$]/$N$. The relationship between $Q$ and the joint cross-entropy $H(X, Y|\theta)$ is

$$\begin{aligned} Q&=\frac{1}{N}\log P(\mathbf{X},\mathbf{Y}|\theta)=\frac{1}{N}\log\prod_j[P(y_j)^{N_j}P(x(1),x(2),...,x(N_j)|\theta_j)] \\ &=\frac{1}{N}\log[\prod_j[P(y_j)^{N_j}\prod_i P(x_i|\theta_j)]=\sum_j\sum_i P(x_i,y_j)\log P(x_i,y_{\theta j})= -H(X,Y|\theta). \end{aligned} \tag{7}$$

Therefore, $Q$ is a negative joint cross-entropy $-H(X, Y|\theta)$.

*2.2. Analyzing the Q-Proof of the EM Algorithm*

Assume that $n$ Gaussian functions (true models) are:

$$P^*(x|y_j)=P(x|\ \theta_j^*)=K_j\exp[-(x-\mu_{j^*})^2/(2\sigma_j^{*2})],\quad j=1, 2, \ldots, n, \tag{8}$$

where $K_j$ is a normalizing coefficient; $\mu_j$ is the mean; $\sigma_j$ is the standard deviation; * means that the distribution or the parameter is true. In the following, we assume $n$=2.

For theoretical convenience, we assume that $P(x)$ is a smooth distribution function ($|P(x_{i+1})-P(x_i)|$ is tiny for any $i$) produced from a true mixture model so that the minimum of relative entropy $H(P||P_\theta)$ is zero. Practically, $P(x)$ may come from a sample with a certain size and is unsmooth. $H(P||P_\theta)\rightarrow 0$ used as the stop condition is better than $\Delta L(\mathbf{X}|\theta)\rightarrow 0$ because using the former, we can easily distinguish local convergence and global convergence. If $P(x)$ is unsmooth, we should replace 0 with a tiny number. The smaller the sample size is, the bigger this number is.

Assume that we only know $P(x)$ and $n$=2 without knowing the true model parameters and the true mixing proportions. We can only guess the distribution:

$$P_\theta(x)=P(y_1)P(x|\theta_1)+P(y_2)P(x|\theta_2). \tag{9}$$

If two distributions are close to each other so that the relative entropy $H(P||P_\theta)$ is close to 0, such as less than 0.001 or 0.01 bit, then we can say that our guess is right. Therefore, our task is to change $P(y)$ and $\theta$ to maximize log-likelihood $L$ or to minimize relative entropy $H(P||P_\theta)$.

In the M-step, the relation between $Q$ and $L$ is

$$\begin{aligned}
L &= \frac{1}{N}\log P(\mathbf{x}|\theta) = \sum_i P(x_i)\log P_\theta(x_i) \\
&= \sum_i\sum_j P(x_i)P(y_i|x_i,\theta^{(t)})\log[P(x_i,y_j|\theta)/(P(x_i,y_j|\theta)/P_\theta(x_i))] \\
&= Q - \sum_i\sum_j P(x_i)P(y_i|x_i,\theta^{(t)})\log P(y_i|x_i|\theta) = Q + H(Y|X,\theta) \\
&\geq Q - \sum_i\sum_j P(x_i)P(y_i|x_i,\theta^{(t)})\log P(y_j|x_i,\theta^{(t)}) = Q + H,
\end{aligned} \tag{10}$$

where $\theta^{(t)}$ denotes the $\theta$ before the M-step; $P(y_j|x,\theta^{(t)})$ is obtained from the last E-step; $H=H(Y|X)$ is a Shannon conditional entropy. The "≥" exists because of the Property 1 of the cross-entropy.

Since $L \geq Q+H$, we can optimize a mixture model by maximizing $Q$. Using the cross-entropy method, we do not need Jensen's Inequality to explain the EM algorithm. Two steps in the EM algorithm are:

**E-step:** Write the conditional probability functions:

$$\begin{aligned}
&P(y_j|x)=P(\theta_j|x) = P(y_j)P(x|\theta_j)/P_\theta(x),\ j=1,2,\ldots,n; \\
&P_\theta(x) = \sum_j P(y_j)P(x|\theta_j).
\end{aligned} \tag{11}$$

**M-step**: Maximize $Q$ by improving $P(y)$ and $\theta$ in log( ). If we cannot improve $Q$ further, then end the iteration; otherwise, go to the E-step.

We can separate the M-step into the M1-step for new $P(y)$ and the M2-step for new $P(x|\theta_j)$, j=1,2,…,$n$.

To distinguish $L$, $Q$, and $H(Y|X,\theta)$ in different stages, we use parameters $\theta^{(t)}$ (before the M-step) and $\theta^{(t+1)}$ (after the M-step) to express them. For example, $Q(\theta^{(t+1)}|\theta^{(t)})$ means that the parameter in $P(y|x)$ is $\theta^{(t)}$, and that in $P(x|\theta_j)$ is $\theta^{(t+1)}$. Hence, we have

- $L(\mathbf{X}|\theta^{(t)})= L(\theta^{(t)}|\theta^{(t)})= Q(\theta^{(t)}|\theta^{(t)})+H(\theta^{(t)}|\theta^{(t)})$ before the M-step,
- $L(\theta^{(t+1)}|\theta^{(t)})= Q(\theta^{(t+1)}|\theta^{(t)})+H(\theta^{(t+1)}|\theta^{(t)}) \geq Q(\theta^{(t+1)}|\theta^{(t)})+H(\theta^{(t)}|\theta^{(t)})$ after the M-step, and
- $L(\mathbf{X}|\theta^{(t+1)})=L(\theta^{(t+1)}|\theta^{(t+1)})= Q(\theta^{(t+1)}|\theta^{(t+1)})+H(\theta^{(t+1)}|\theta^{(t+1)})$ after the E-step.

After the E-step, $\theta^{(t)}$ becomes $\theta^{(t+1)}$, $P(y_j|x)$ is $P(\theta_j^{(t+1)}|x)$, and $H(Y|X,\theta)$ becomes Shannon's conditional entropy:

$$H(\theta^{(t+1)} \mid \theta_j^{(t+1)}) = H^{(t+1)}(Y \mid X) = -\sum_i \sum_j P(x_i) P(\theta_j^{(t+1)} \mid x_i) \log P(\theta_j^{(t+1)} \mid x_i). \quad (12)$$

In the Shannon information theory, transition probability matrix $P(Y|X)$ is called the channel; therefore, we call conditional probability distribution $P(y|x)$ obtained from the E-step the Shannon channel.

Since $H(\theta^{(t+1)}|\theta^{(t)}) \geq H(\theta^{(t)}|\theta^{(t)})$ according to the Property 1 of the cross-entropy, we have

$$L(\theta^{(t+1)}|\theta^{(t)}) - L(\theta^{(t)}|\theta^{(t)}) \geq Q(\theta^{(t+1)}|\theta^{(t)}) - Q(\theta^{(t)}|\theta^{(t)}) \geq 0 \text{ (after the M-step)}. \quad (13)$$

It is this inequality or a similar one that is used in the *Q-Proof*. However, we can only use this inequality to prove that the M-step increases $Q$ and $L$. $Q$ and $L$ also change in the E-step. To confirm that the E-step or every iteration increases $Q$ and $L$, we also need to prove

$$L(\theta^{(t+1)}|\theta^{(t+1)}) - L(\theta^{(t+1)}|\theta^{(t)}) \geq Q(\theta^{(t+1)}|\theta^{(t+1)}) - Q(\theta^{(t+1)}|\theta^{(t)}) \geq 0 \text{ (after the E-step)}. \quad (14)$$

However, $Q(\theta^{(t+1)}|\theta^{(t+1)})-Q(\theta^{(t+1)}|\theta^{(t)})>0$ in Equation (14) has never been proved. It is impossible to prove this inequality because the E-step lets $P(y|x)$ approach $P(y|x, \theta)$, which may decrease $Q$ (according to the Property 2 of cross-entropy). Sections 3.1.1 and 3.1.2 provide two counterexamples against Equation (14).

Now we can easily explain why Ueda and Nakano's proof [10] for that the E-step maximizes $Q$ is incorrect. In that proof, the author explains that we can obtain Equation (11) (in this paper) in the E-step by letting $\partial Q(\theta|\theta^{(t)})/\partial\theta=0$ (see Equations (5) and (6) in [10]). However, $\partial Q(\theta|\theta^{(t)})/\partial\theta=0$ is the condition under which the M-step maximizes $Q(\theta|\theta^{(t)})$ and $L(\theta|\theta^{(t)})$. By changing $\theta$ so that $\partial Q(\theta;\theta^{(t)})/\partial\theta=0$, we cannot obtain Equation (11) for the E-step.

The following two proofs are a little different from the *Q-Proof*.

**The Weak *Q*-Proof:**

One may say that the E-step can improve $Q$ instead of increasing $Q$; if we can change $\theta$ or $Q$ so that $L$ is non-decreasing, the convergence proof also holds. He is right. But we also need to prove that $L$ is non-decreasing no matter if $Q$ increases in the E-step. We call this proof the weak *Q-Proof*.

The difficulty of this proof is that after the E-step, $H$ also changes from $H(\theta^{(t)}|\theta^{(t)})$ to $H(\theta^{(t+1)}|\theta^{(t+1)})$. Using equality $L= Q+H(Y|X,\theta)$ and inequality $L \geq Q+H$ in Equation (10), we can easily find the problem.

The author was told that Davison [38] proposed proof that directly proves the E-step increases $L$. However, we can find that Davison's proof is similar to Ueda and Nakano's proof in [10]. The Inequality (5.39) in [38], similar to Inequality (13) in this paper, is used for both steps. This proof is also incorrect. To prove that the E-step increases $L$, we need another method.

The author examined the proof in Stanford University CS229 [29] provided by Tengyu Ma and Andrew Ng. There is a similar problem. Because [29] is famous and has many readers, this paper offers Appendix B to clarify its issue.

Only Beal [13] proposed such proof with the variational method for the E-step. His variational proof is also a weak $Q$-Proof. This proof considers that $H$ changes with $P(y|x)$ in the E-step.

**The Simple *Q*-Proof:**

In some proofs, authors do not mention if $Q$ and $L$ increase or decrease together, neither mention if the E-step changes $L$. It seems that $L$ does not change in the E-step. If so, it is much easy to prove the convergence of the EM algorithm. We call the proof that affirms that E-step does not change $L$ the simple $Q$-Proof.

If this proof is correct, Equation (11) for the E-step is unnecessary, and any change of $P(y|x)$ can ensure that the next M-step increases $L$. For example, we may always use $P(y_1)=P(y_2)=0.5$ or arbitrary $P(y)$ to update $P(y|x)$ in the E-step. According to the simple $Q$-Proof, $L$ can continuously increase after every iteration. However, this result is peculiar.

What is the problem? This proof neglects $H$'s change. Anyway, $H$ must change from $H(\theta^{(t)}|\theta^{(t)})$ to $H(\theta^{(t+1)}|\theta^{(t+1)})$ with $P(y|x)$ no matter if it changes in the E-step. Once it changes, it must influence $P_\theta(x)$ and $L$. Therefore, we need to prove that this influence is positive.

In brief, the $Q$-Proof and the simple $Q$-Proof are incorrect; the weak $Q$-Proof and the variational proof may be correct.

In Neal and Hinton's MM algorithm [17], the objective function $Q$ is replaced with $F=Q+H(Y)$, where $H(Y)$ is the Shannon entropy of $Y$. The MM algorithm maximizes $F$ in both M-step and E-step so that the convergence is faster. However, it has not been proved that $F$ and $L$ are positively correlated. We can also find counterexamples (see Figure 2 (c), Figure 5 (c)).

## 3. The Convergence Analysis of the EM Algorithm

### *3.1. Evidence that Reveals Mistakes in The Q-L Synchronization Theory*

#### 3.1.1. A Counterexample with $Q \geq Q^*$ against the Two Affirmations

To prove that the above Affirmation I and Affirmation II are the two theoretical mistakes in the $Q$-$L$ synchronization theory that bases the DAEM algorithm, let us see a counterexample.

A true mixing proportion distribution $P^*(y)$ and a true conditional probability distribution $P^*(x|y)$ ascertain the joint probability distribution $P^*(x, y) = P^*(y)P^*(x|y)$. The corresponding joint entropy is

$$H^*(X,Y) = -\sum_j \sum_i P^*(x_i, y_j) \log P^*(x_i, y_j) = -Q^*. \qquad (15)$$

Now we show that $Q$ might be greater than $Q^*$.

***Example 1.*** *$U$={1, 2, 3, …, 150}; a true model is ($\mu_1^*$, $\mu_2^*$, $\sigma_1^*$, $\sigma_2^*$, $P^*(y_1)$)=(65, 95, 15, 15, 0.5). Suppose that the guessed ratios and parameters are $P(y_1)$=0.5, $\mu_1$= $\mu_1^*$, $\mu_2$=$\mu_2^*$, and $\sigma_1$= $\sigma_2$=$\sigma$. Figure 1 shows that $Q$ and $L$ change with $\sigma$.*

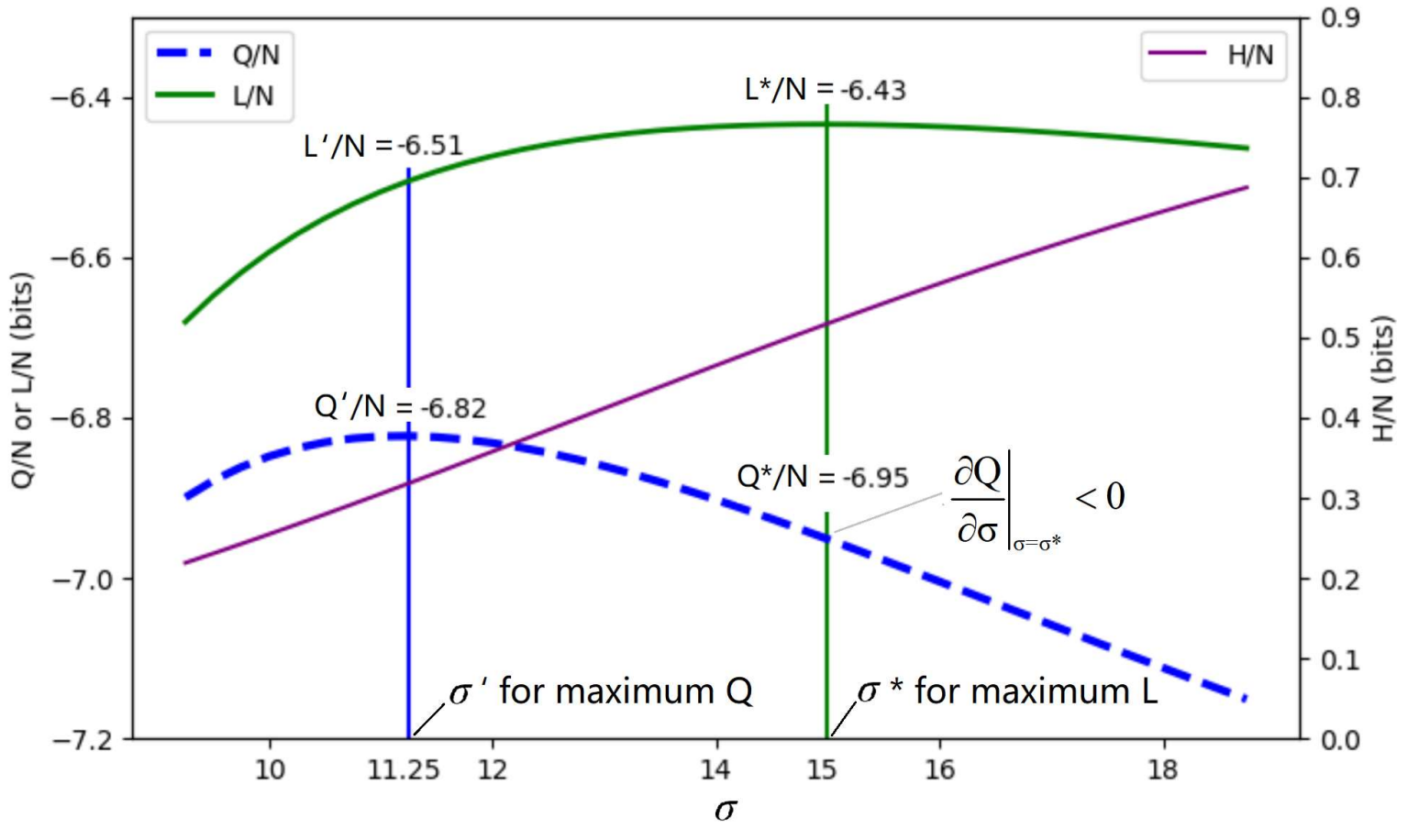

Figure 1. $Q$ and $L$ change with $\sigma$ against the two affirmations about the EM algorithm. The convergent $Q$=$Q^*$ is not the maximal $Q$=$Q'$. One can find the source files in Python 3.6 for Figures 1, 2, 5, 7-12 in Supplemental Materials.

In this example, when $\sigma$ changes from $\sigma'$ =11.25 to $\sigma^*$=15, $Q$ decreases from its global maximum $Q'$ = -6.82 bits to $Q^*$= -6.95 bits; whereas $L$ increases from $L(\sigma')$= -6.51 bits to its global maximum $L(\sigma^*)$= -6.43 bits. Therefore, Affirmation I is wrong. We can use the EM algorithm to solve the above example with an initial $\sigma$=11.25. Figure 2 (c) shows that $Q$ changes with $H(P||P_\theta)$ in the EM algorithm. The sample size is 50000.

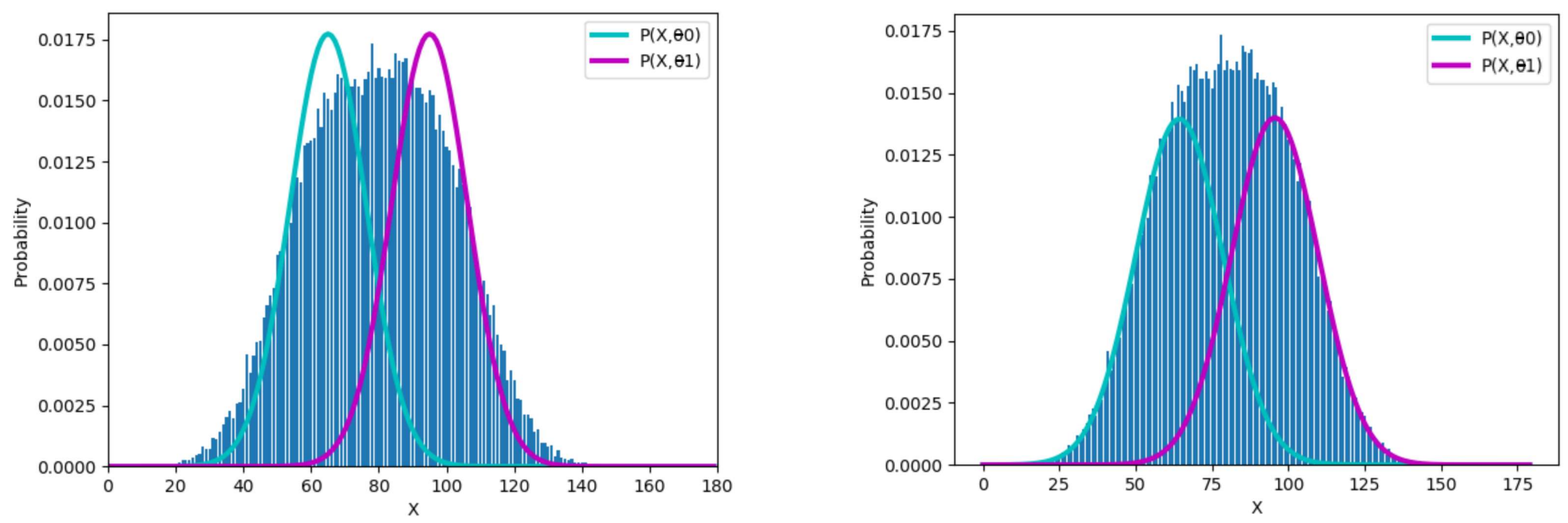

(a) Initialization with larger $Q$= -6.82 bits. (b) Valid convergence with smaller $Q^*$= -6.95 bits.

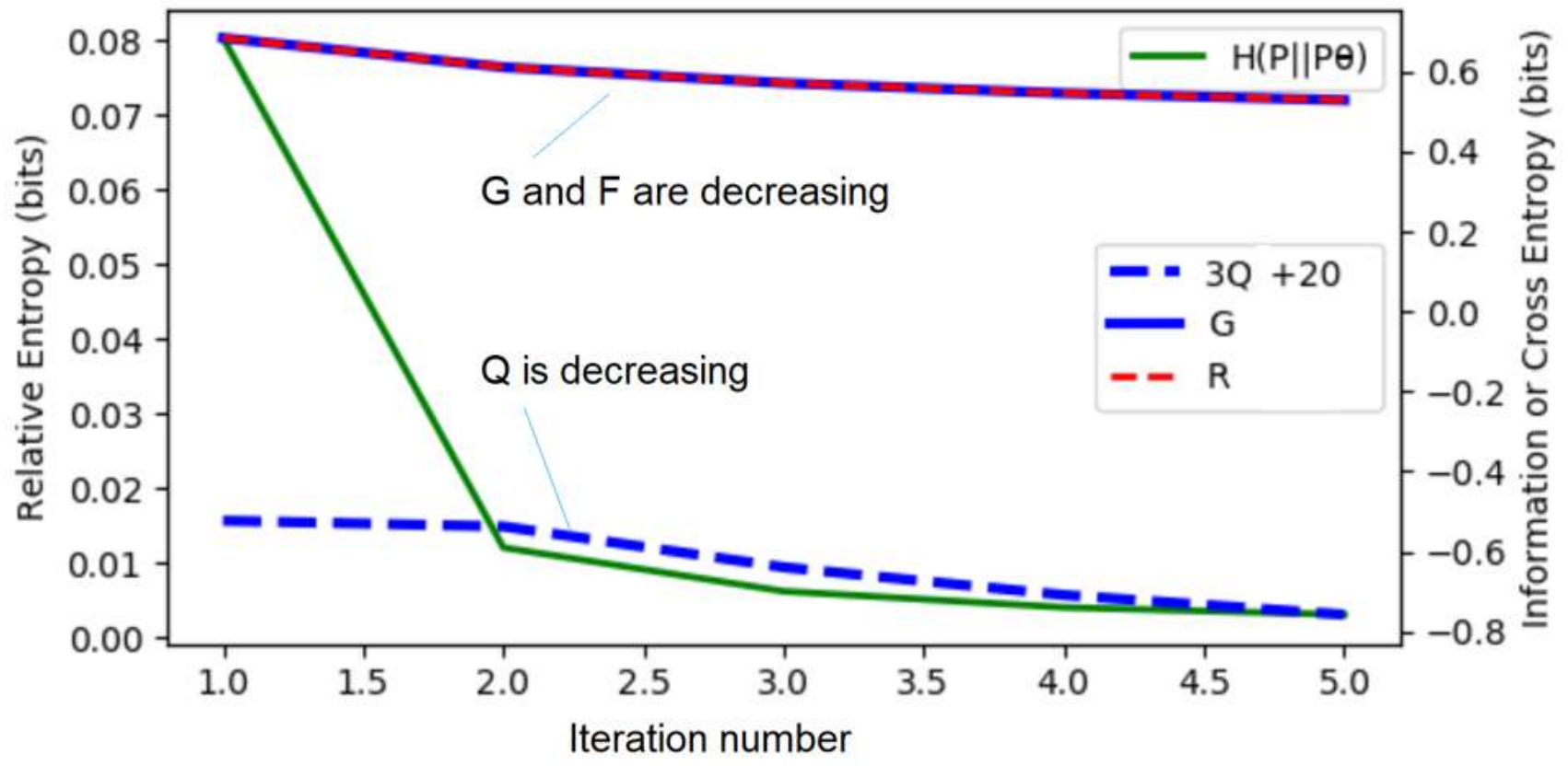

(c) $Q$ and $F$ decrease as $H(P||P_\theta)$ decreases or $L$=$L(\mathbf{X}|\theta)$ increases.

Figure 2. A counterexample against the two affirmations about the EM algorithm. $Q$ decreases after every iteration in the EM algorithm. $G$ and $F$ also decrease with $H(P||P_\theta)$ against the MM algorithm.

Figure 2 indicates that the E-step can decrease $Q$. When $Q$>$Q^*$, $Q$ can converge to $Q^*$ only because the E-step can reduce $Q$. We can find that the EM algorithm can converge in some cases not because the E-step is non-decreasing but because the relative entropy $H(P||P_\theta)$ is decreasing. That means that, in the $Q$-Proof, the mistake in Affirmation II covers up the error in Affirmation I.

### 3.1.2. The Mathematical Analysis about that $Q$ Might Be Greater than $Q^*$

Further, the author found that for any true model parameter set $\theta^*$ and the corresponding $Q^*=Q(\theta^*)= Q(\mu_1^*, \mu_1^*, \sigma_1^*, \sigma_2^*, P^*(y_1))$, we could always find a ratio $r$ between 0.5 and 1, such as $r$=0.75, so that $Q(\theta')= Q(\mu_1^*, \mu_1^*, r\sigma_1^*, r\sigma_2^*, P^*(y_1))>Q(\theta^*)$. To prove this conclusion, we need to prove $\left.\frac{\partial Q}{\partial \sigma_j}\right|_{\sigma_j=\sigma_j^*}<0$. This proof is not easy; nevertheless, we can prove that this conclusion is tenable for some symmetrical mixture models. In Example 1, we further suppose that $P(y_1)=P^*(y_1)=0.5$, $\mu_1=\mu_1^*$, $\mu_2=\mu_2^*$, and $\sigma_1=\sigma_2=\sigma$. Hence, $Q$, $L$, and $H$ only change with $\sigma$. We have $Q=Q(\sigma)$, $L=L(\sigma)$, and $H=H(\sigma)$. $L(\sigma^*)$ is the maximum of $L$. Now, we can prove $\left.\frac{dQ}{d\sigma}\right|_{\sigma=\sigma^*}<0$.

**Proof**: According to Equation (12) for the E-step,

$$P(y_1|x)=\frac{P(y_1)e^{-(X-\mu_1)^2/(2\sigma^2)}}{P(y_1)e^{-(X-\mu_1)^2/(2\sigma^2)}+P(y_2)e^{-(X-\mu_2)^2/(2\sigma^2)}}=\frac{1}{1+e^{b/\sigma^2}}. \quad (16)$$

where $b$=[($\mu_2$-$\mu_1$)$x$-($\mu_1^2$-$\mu_2^2$)]/2. $P(y_1|x)$ is a Logistic function. We can prove d$H$/d$\sigma$>0 (see Appendix C for the detailed proof).

Since d$L$/d$\sigma$=0 as $\sigma$=$\sigma$* is the condition that maximizes $L$, and $Q$=$L$-$H$, we have

$$\left.\frac{dQ}{d\sigma}\right|_{\sigma=\sigma^*}=\left.\frac{dL}{d\sigma}\right|_{\sigma=\sigma^*}-\left.\frac{dH}{d\sigma}\right|_{\sigma=\sigma^*}=0-\left.\frac{dH}{d\sigma}\right|_{\sigma=\sigma^*}<0. \tag{17}$$

**QED.**

When $\sigma$ increases, the Logistic function becomes flatter, and hence Y's posterior entropy $H(Y|X)$ is greater. The author has tested many true models' parameters. For every $\theta$*, we can always find $\theta$' so that $Q(\theta')$ is bigger than $Q(\theta^*)$ by replacing $\sigma_j$* with $\sigma_j' = r\sigma_j$* ($r$ is about 0.75).

3.2. The Shannon Channel $P(Y|X)$ from the E-step for the Expectation is Abnormal

Shannon calls the conditional probability matrix $P(Y|X)$ as the channel. A Shannon's channel consists of a group of transition functions $P(y_j|x)$ (j=1,2,…), where $y_j$ is a constant, and $x$ is a variable. $P(y|x)$ that E-step uses for the expectation is a Shannon channel. A good Shannon's channel $P(y|x)$ should makes

$$P(y_j)=\sum_i P(x_i)P(y_j|x_i)=\sum_i P(x_i|y)P(y_j)=P(y_j),\ j=1,2,...,n. \tag{18}$$

However, in the E-step, for given $P(x)$, $P(y)$, and $\theta$, we have a new $P(y)$ denoted by $P^{+1}(y)$:

$$P^{+1}(y_j)=\sum_i P(x_i)P(y_j|x_i)=\sum_i P(x_i)P(x_i|\theta_j)P(y_j)/P_\theta(x_i). \tag{19}$$

Generally, there is $P^{+1}(y) \neq P(y)$. Therefore, the Shannon channel from the E-step is abnormal. Even if $P^{+1}(y) \neq P(y)$ is not a mistake, at least it is improper. To provide a proper Shannon channel, we need to find the unique $P(y)$ that matches $P(x)$ and $\theta$ so that $P^{+1}(y)=P(y)$. In the rate-distortion theory, there is a similar task. In the CM-EM algorithm, an iteration method is used for $P^{+1}(y)=P(y)$ (see Section 4.1).

*3.3. Why Does the Typical Local Convergence Happen?*

According to the *Q-L* synchronization theory, Ueda and Nakano [10] proposed the DAEM algorithm. They conclude that when some initial parameters result in locally maximal $Q$, $L$'s local or invalid convergence is inevitable; we may use the Deterministic Annealing (DA) method, which increases standard deviations, to avoid locally maximal $Q$ and $L$.

Example 2 is provided in [10] and verified by Marin *et al.* in [11].

***Example 2.*** *A mixture model has two Gaussian components. The true model is ($\mu_1$*, $\mu_2$*, $\sigma_1$*, $\sigma_2$*, $P^*(y_1)$)= (0, 2.5, 1, 1, 0.7) (see Figure 3 from [11]). Marin et al. show that among the five initial points on the $\mu_1$-$\mu_2$ plane, using the EM algorithm, only two points validly converge to ($\mu_1$*, $\mu_2$*); the other three points invalidly converge to a point near ($\mu_1$, $\mu_2$)=(1.5, -0.5).*

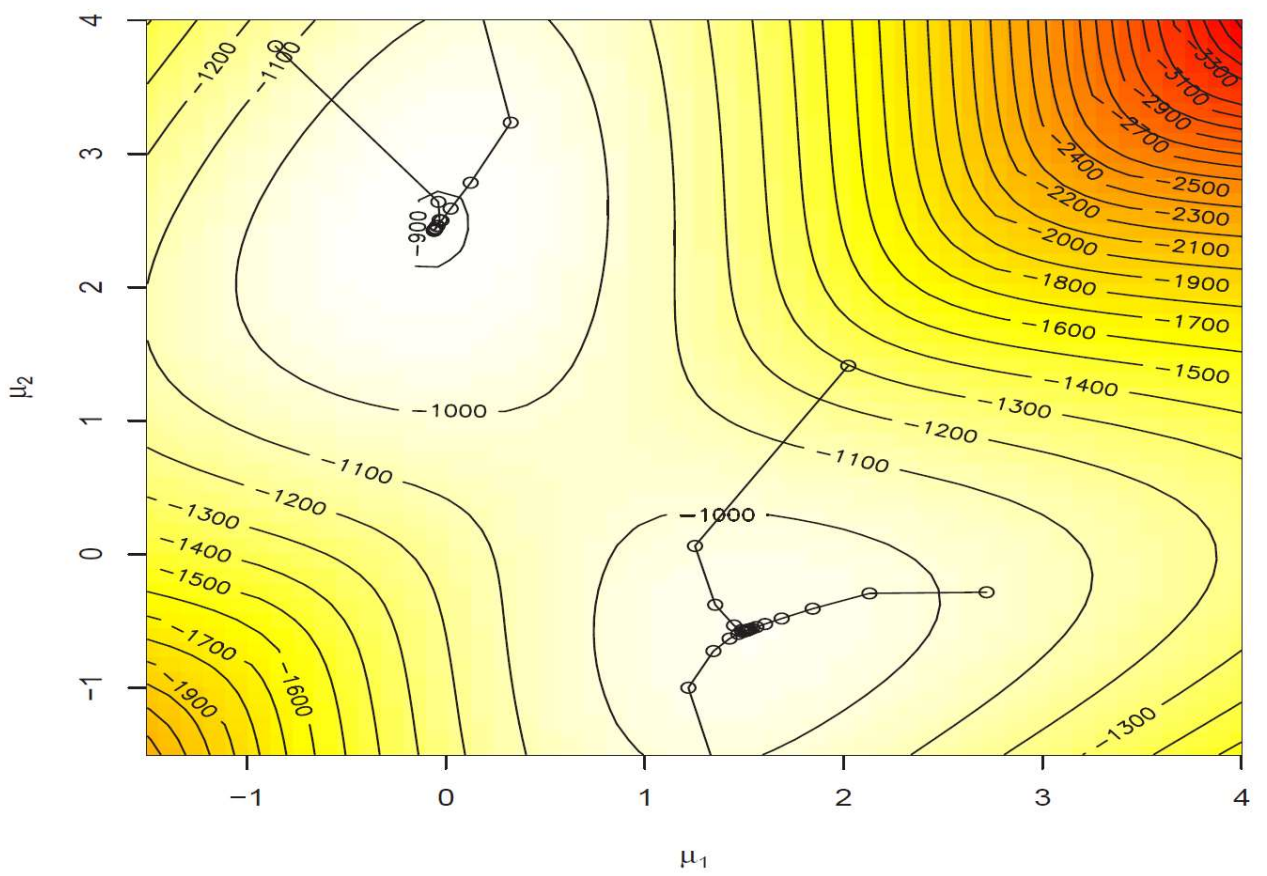

Figure 3. Trajectories of five runs of the EM algorithm on the log-likelihood surface (Thanks to the authors of [11] for the kind permission). The authors assume that $\sigma_1$, $\sigma_2$, and $P(y_1)$ are fixed and neglect that two mixture models with symmetrical parameters are true at the same time.

In [11] and Figure 3, contour lines represent the log-likelihood $L$'s distribution. In [10], contour lines represent $F$'s distribution for annealing parameter $\beta$=1, where $F$ means free energy and is equal to -$L$. Both authors believe that point (1.5, -0.5) has locally maximal $Q$ and $L$. They could draw the $L$'s distribution because they had fixed parameters $\sigma_1$, $\sigma_2$, and $P(y_1)$. However, in the real EM algorithm, $L$'s distribution must change with the iterative process and every parameter.

Later, to avoid the local convergence and boundary convergence, Ueda et al. proposes the Split and Merge EM (SMEM) algorithm [13]; Marin *et al.* use the Population Monte Carlo (PMC) algorithm [11].

After inspecting the EM algorithm's behaviors for Example 2, with different sample sizes and various initial points on the $\mu_1$-$\mu_2$ plane, the author reached different conclusions. Figure 4 is used to explain these conclusions. The ($\mu_1$, $\mu_2$) in Figure 3 becomes (10$\mu_1$+100, 10$\mu_2$+100) in Figure 4.

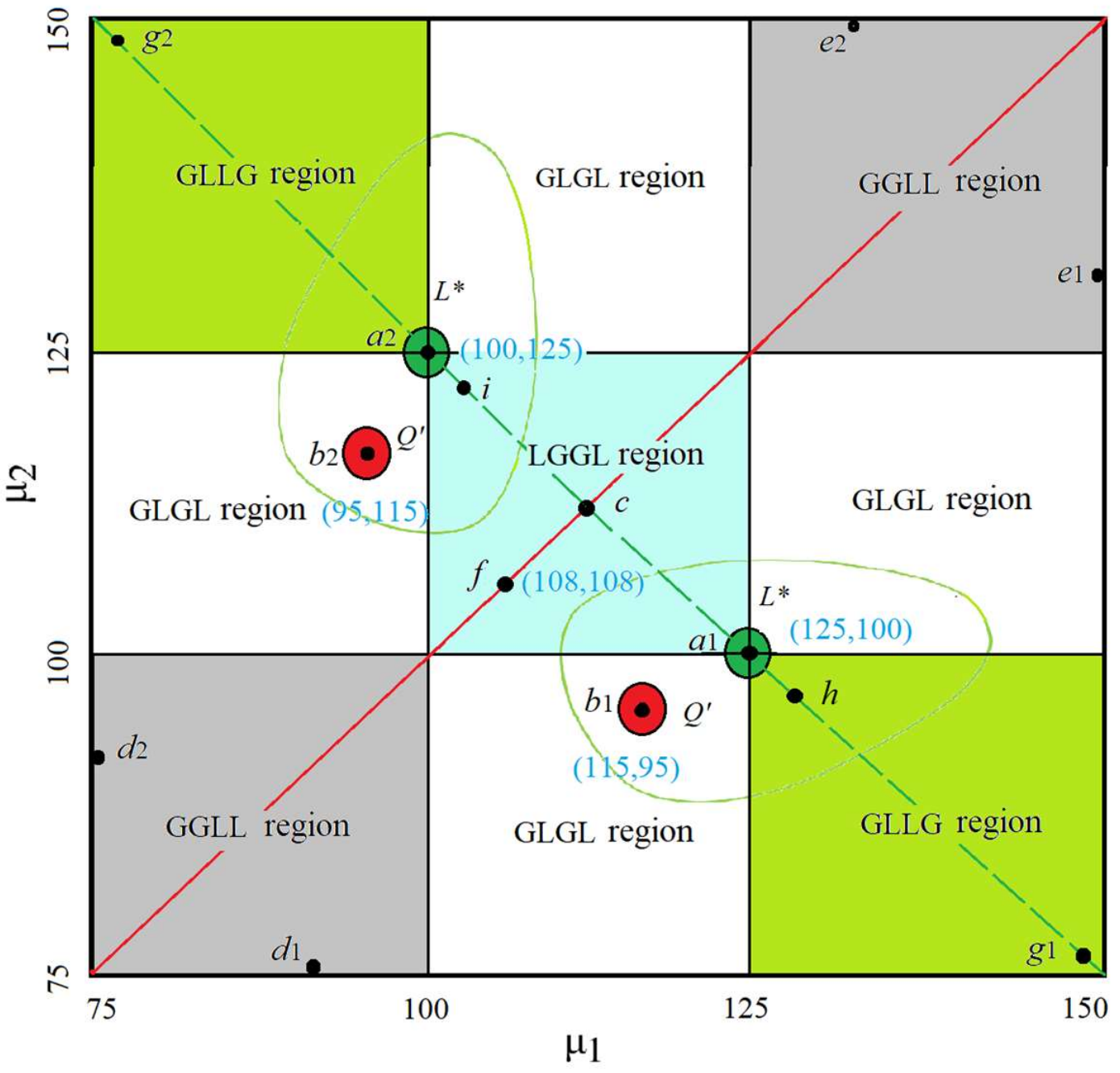

Figure 4. Symmetrical $\mu_1$-$\mu_2$ plane with two pairs of locally and globally convergent points. The four kinds of regions have different convergent difficulties. Green regions are the easiest; gray regions are the hardest.

These conclusions are related to

- **Symmetry**:

a) If model ($\mu_1$, $\mu_2$, $\sigma_1$, $\sigma_2$, $P(y_1)$) = (100, 125, 10, 10, 0.7) produces sampling distribution $P(x)$, then model ($\mu_1$, $\mu_2$, $\sigma_1$, $\sigma_2$, $P(y_1)$) = (125, 100, 10, 10, 0.3) (two components' parameters are exchanged) must produce the same $P(x)$. We can consider both models as true models. In other words, there are two points, $a_1$ and $a_2$ in Figure 4, where $L$ is globally maximal, denoted by $L^*$; there are also two points $b_1$ and $b_2$, where $Q$ is locally maximal, denoted by $Q'$. The two areas divided by the 45° line (red or solid line) are axisymmetric, with the 45° line as the symmetry axis. This line is like a deep ditch with smaller $Q$ and $L$; the EM algorithm cannot cross this deep ditch if $\sigma_1$ and $\sigma_2$ have no big difference. If the initial point is in this deep ditch and $\sigma_1$=$\sigma_2$, the EM algorithm must invalidly converge to point $f$, which is the deep ditch's shallowest point.

b) In Figure 3 above and Figure 2 in [10], the contour lines are asymmetrical because the authors [10,11] always use $P^*(y_1)$=0.7 for both sides. However, for the lower side, we should use $P^*(y_1)$=0.3 to draw the contour lines because when the EM algorithm converges to $b_1$ or $a_1$, $P(y_2)$ is close to 0.7 rather than $P(y_1)$. The authors of [10,11] probably think that every convergence to points below the 45° line is local convergence, and every convergence to points above the 45° line is global convergence. Ueda and Nakano [10] provide two bad initial points on the lower side, with which the EM algorithm converges to $b_1$, whereas the DAEM algorithm converges to $a_2$. In this example, the DAEM algorithm makes great efforts to cross the deep ditch (the 45° line). However, we need not seek far and neglect what lies close at hand. The $a_1$ is as good as $a_2$. Because the convergence to $a_1$ looks the same as the convergence to $a_2$ (see Figure 5(b)), someone might mistake $a_1$ for $a_2$. Because $a_1$ and b1 are similar, someone might mistake $a_1$ for $b_1$.

c) $P^*(y_1)\neq P^*(y_2)$ is the cause of the existence of $b_1$, $b_2$, and $f$. When $P^*(y_1)$ changes from 0.7 to 0.5, points $b_1$, $b_2$, and $f$ gradually disappear or move to $a_1$, $a_2$, and $c$, respectively. Then Figure 4 becomes centrosymmetric.

- **Local convergence and global convergence**: Local convergence happens only when sample sizes are too small, such as $N$=1000, and the relative positions of $\mu_1$, $\mu_2$, $\mu_1^*$, and $\mu_2^*$ are improper. Other points between $b_1$ and (120, 98) or between $b_2$ and (98, 120) are also possibly convergent points, which means that the EM algorithm is easily stuck due to fewer sample points. If samples are big enough, such as $N$=50000, the EM algorithm can let $L$ converge to $L^*$ from any initial point. Figure 5 shows a run with a very bad initial point $d_2$=(80,95).
- **Relative positions and convergence difficulties**: There are 4!=24 permutations of $\mu_1$, $\mu_2$, $\mu_1^*$, and $\mu_2^*$. Suppose some symmetrical permutations are regarded as the same. In that case, all permutations (or nine regions in Figure 4) can be classified into four kinds represented by four labels: GLLG, LGGL, GLGL, and GGLL. If we put the initial ($\mu_1$, $\mu_2$) in these regions, convergence difficulties increase in turn. When $P(y_1)$=$P(y_2)$=0.5, Figure 4 is centrosymmetric; all regions with the same label are equally easy for global convergence. If $P(y_1)\neq P(y_2)$, then two regions with the same label might be unequally difficult for global convergence. In Figure 7, iteration numbers that are put in some typical locations represent the convergence difficulties of those locations. In short, (1) points close to the 135° line in GLLG or LGGL regions, such as $g_1$, $g_2$, $h$, and $i$, are the best points; (2) points close to the 45° line in GGLL or LGGL regions are worse points; (3) points in or behind a $Q$-larger area, such as areas close to $b_1$, $b_2$, $d_1$, and $d_2$, are worst points.
- **About the annealing operation**: When samples are big enough, it is not necessary to increase $\sigma_1$ and $\sigma_2$, such as using $\sigma_1$=$\sigma_2$=20, which means that the annealing operation is used. When samples are small, adjusting the initial relative position or ($\mu_1$, $\mu_2$) is more effective than changing standard deviations.
- ***Q* changes with *L*:** The EM algorithm can pass through $Q$-larger and $Q$-smaller areas to reach $L^*$ (see Figure 5); $Q$ and $L$ are not positively correlated. Figure 5 (a) shows that the two components with maximal $Q$. Figure 5 (b) shows the two components with maximal $L$. Figure 5 (c) shows the iteration process of the EM algorithm for Example 2.

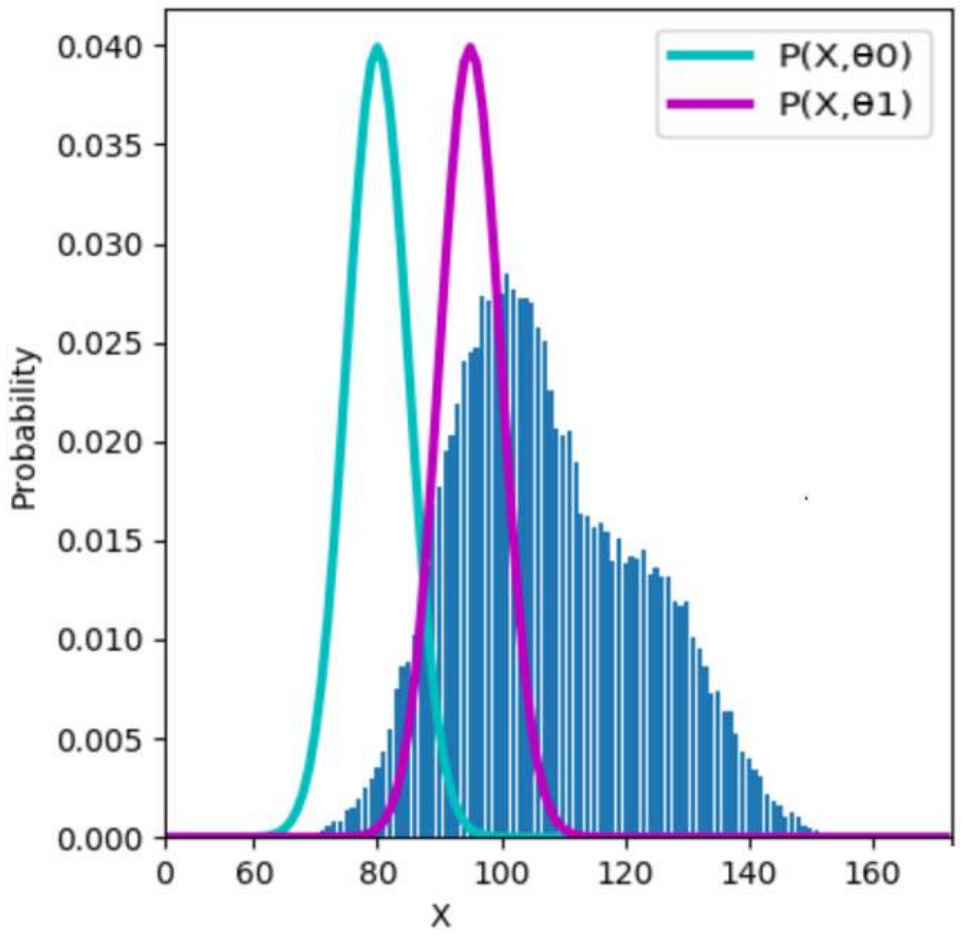

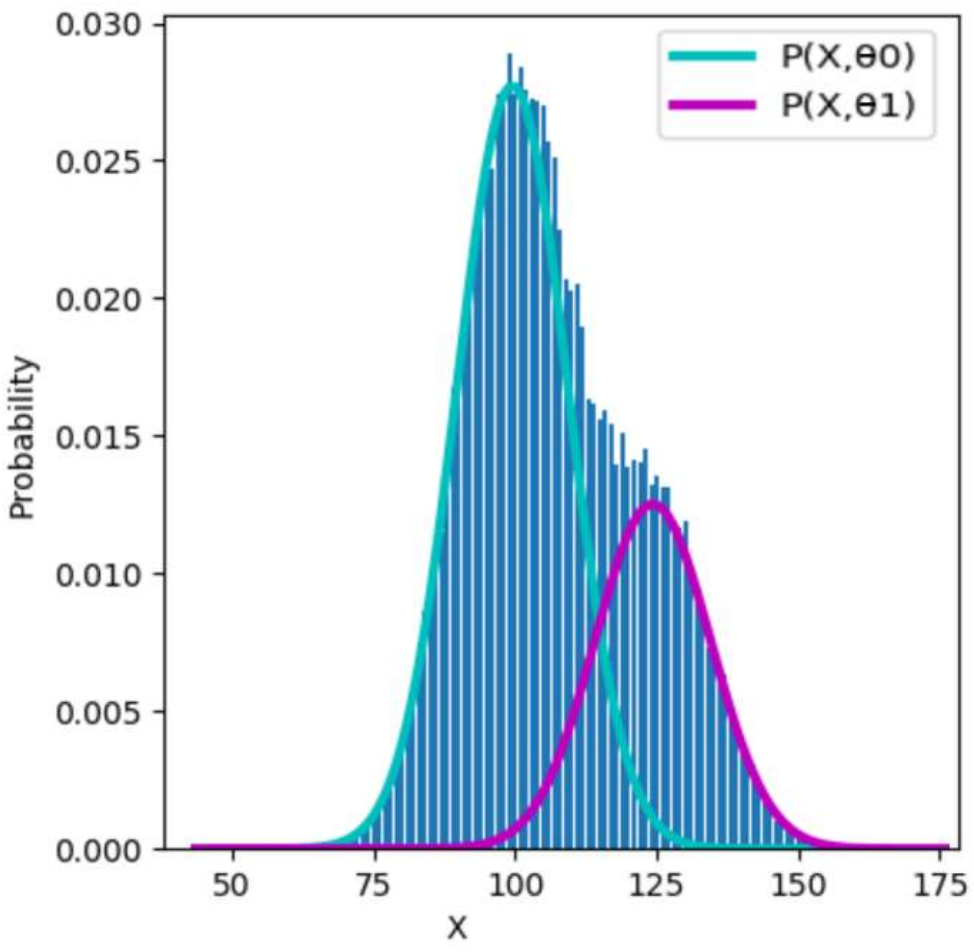

(a) Initial components with ($\mu_1$, $\mu_2$)=(80,95). (b) Globally convergent two components.

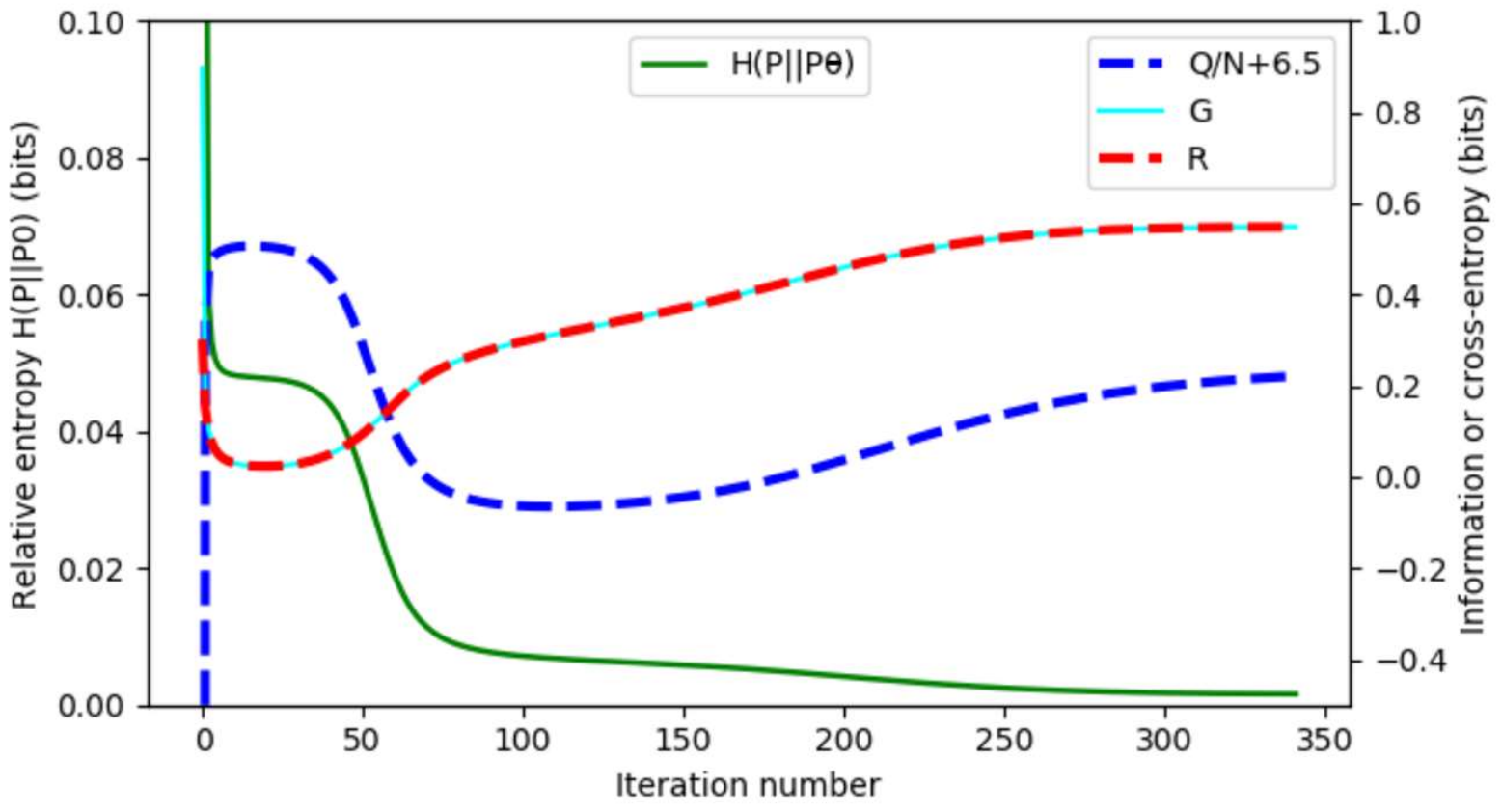

(c) $Q$ and $H(P||P_\theta)$ changes with iterations (initialization: ($\mu_1$, $\mu_2$, $\sigma_1$, $\sigma_2$, $P(y_1)$)= (80, 95, 5, 5, 0.5)).

**Figure 5**. The EM algorithm for Example 2 ($N$=50000) is against the $Q$-$L$ Synchronization Theory. The EM algorithm needs about 340 iterations. The E3M algorithm needs about 240 iterations (see Figure 7).

In Figure 5 (c), the initial $\sigma_1$ and $\sigma_2$ are 5, so that $Q'$ is globally maximal. Although $Q'$ is the globally maximal $Q$, it cannot stop ($\mu_1$, $\mu_2$) to reach $a_2$ with a smaller $Q^*$. The $Q$-$L$ synchronization is falsified again.

### *3.4. Explaining Convergence Difficulties Using Marriage Competition*

Why are some initial points on the $\mu_1$-$\mu_2$ plane easy or difficult to converge to $a_1$ and $a_2$ with $L^*$? We use a metaphor to explain the reason.

Imagine that two ladies (denoted by L and L) and two gentlemen (denoted by G and G) live along a road with relative positions like these: "__G__G__L__L__", "__G__L__G__L__", and "__G____LG__L_". The valid or global convergence means that one G marries with one L. Because of the unfair competition, the left G is not easy to attract the left L, which is also attracted by the right G. If they have cars so that distances are no obstructions, then valid convergence is a little easy. An L is like a sub-sample, and a G is like a component. The Deterministic Annealing method seemly provides cars to enlarge two gentlemen's scopes of activities. However, it is more effective to adjust two gentlemen's relative positions. Assume that distances between any two adjacent people are equal. Their relative positions can be simply represented by four labels: GLLG, LGGL, GLGL, and GGLL, which reflect different convergence difficulties.

Distances between them also affect convergence difficulties. For example, "__G____LG__L_" is more difficult for the global convergence than "__G__L__GL_" because "LG" in the former are not easy to separate. If we consider that mixture proportions are different, and a sub-sample with a larger mixture proportion is like a more attractive lady, which is denoted by **L**, then "__G___**L**G__L_" is more difficult than "__G___LG__L_" for the global convergence. The most difficult relative position is "__G__G___ **L** ____L_" (such as $d_1$ and $d_2$ in Figure 4), which will become "__G_**L**G__L_", not only because the competition is unfair, but also because it is hard for the right G to give up **L**.

See Figure 5. We call the 135° (green or dashed) line the fair competition line, call the 45° (red or solid) line the absolute equality line, and call the area with locally maximal $Q$ the hard separation area. For the better initial means of components, we should follow three rules:

- Let the initial point ($\mu_1$, $\mu_2$) close to the fair competition line.
- Let the initial point ($\mu_1$, $\mu_2$) apart from the absolute equality line, resulting in "all gentlemen marry with all ladies."
- Do not set the initial point ($\mu_1$, $\mu_2$) in or behind the hard separation areas, such as areas around $b_1$ and $b_2$.

We call the above three rules the FCP.

When two sub-samples in a two-dimensional space have the same means and different standard deviations, perhaps with different correlation coefficients, there is also unfair competition between two components, such as in Figure 10. In these cases, the FCPis helpful but not sufficient. If three components and three sub-samples have relative positions like this: "GGLGLL" or "GGGLLL", for finite-size samples, the left G may be alone, and boundary convergence (a mixing proportion approaches a tiny value or 0) will happen. For avoiding the boundary convergence of more complicated mixture models, the FCP is also useful but not sufficient.

Some researchers think that it is important that initial Gs' barycenter coincides with Ls' barycenter. However, $g_1$ and $g_2$ in Figure 4 indicate that fair competition is more important.

The author's experiments indicate that, in most cases, it is fine that initial mixture proportions are equal, and initial standard deviations are similar and not too small. Too small standard deviations easily result in boundary convergence for both the EM and CM-EM algorithms.

## 4. The Convergence Proof for the EM and CM-EM Algorithms

### *4.1. A Semantic Information Theory with R(G) Function*

The semantic information theory proposed by the author is a generalization of Shannon's information theory. It is based on the P-T probability framework [35].

Shannon [38] calls $P(y_j|x)$ (with certain $y_j$ and variable $x$) the transition probability. A group of transition probability functions forms Shannon's channel: $P(y_j|x)$, $j$=1, 2, …, $n$. Using $P(y_j|x)$, we can make Bayes' prediction:

$$P(x|y_j)=P(x)P(y_j|x)/P(y_j),\ P(y_j)=\sum_i P(x_i)P(y_j|x_i). \tag{20}$$

**Definition 5.** The $\theta_j$ is a predictive sub-model and also a fuzzy subset of $U$; $T(\theta_j|x)$ is the membership function as defined by Zadeh [41]; $y_j$ is a predicate "$x$ is in $\theta_j$." Then $T(\theta_j|x)$ is also the (fuzzy) truth function of $y_j$. A group of truth functions or membership functions form a semantic channel [36]: $T(\theta_j|x)$, $j$=1,2,…,$n$.

The main differences between the transition probability function and the truth function are

- The maximum of $T(\theta_j|x)$ is 1, whereas the maximum of $P(y_j|x)$ is not 1 in general;
- We can obtain $T(\theta_j|x)$ from $P(x|\theta_j)$ (or $P(x|y_j)$) and $P(x)$ without $P(y_j)$, whereas we cannot obtain $P(y_j|x)$ from $P(x|y_j)$ and $P(x)$ without $P(y_j)$.

- The truth function can be explained as the membership function representing a label's extension (denotation).

The truth function $T(\theta_j|x)$ can also be used for Bayes' prediction to produce the likelihood function:

$$P(x|\theta_j)=P(x)T(\theta_j|x)/T(\theta_j),\ T(\theta_j)=\sum_i P(x_i)T(\theta_j|x_i). \tag{21}$$

We call the above formula the semantic Bayes' formula.

We use an example to explain the semantic channel. The $x$ is an age, $y_1$=" $x$ is elderly", and $y_0$=" $x$ is not elderly". A semantic channel is between $X\in\{x|x \text{ is an age}\}$ and $Y\in\{y_0, y_1\}$. A pair of Logistic functions may represent the truth functions of $y_1$ and $y_0$:

$$T(\theta_1|x)=1/[1+e^{-0.2(x-65)}];\ T(\theta_0|x)= e^{-0.2(x-65)}/[1+e^{-0.2(x-65)}]. \tag{22}$$

If $P(x)$ is a population age distribution, then $T(\theta_1)$ is the logical probability of $y_1$.

The semantic information conveyed by $y_j$ about $x$ is defined with log-normalized-likelihood [36]:

$$I(x;\theta_j)=\log\frac{P(x|\theta_j)}{P(x)}=\log\frac{T(\theta_j|x)}{T(\theta_j)}. \tag{23}$$

We also call $I(x;\theta_j)=\log[P(x|\theta_j)/P(x)]$ the predictive information because $P(x|\theta_j)$ is a predictive model.

Averaging $I(x_i;\theta_j)$ for all $i$ and $j$, we have semantic mutual information:

$$\begin{aligned}I(X;Y_\theta)=I(X;Y|\theta)&=\sum_j\sum_i P(y_j)P(x_i|y_j)\log\frac{P(x_i|\theta_j)}{P(x_i)}\\&=\sum_j\sum_i P(x_i)P(y_j|x_i)\log\frac{T(\theta_j|x_i)}{T(\theta_j)}.\end{aligned} \tag{24}$$

Shannon [42] proposed information rate-distortion function $R(D)$ for data compression. Others [43,44] developed the rate-distortion theory. Since the rate-distortion function is equivalent to the information rate-distortion function for memoryless sources, We do not distinguish them in this paper as in Cover and Thomas' textbook [45].

The rate-distortion function with parameter $s$ [43] (P. 32) includes two formulas:

$$\begin{aligned}D(s)&=\sum_i\sum_j d_{ij}P(x_i)P(y_j|x_i),\\R(s)&=sD(s)-\sum_i P(x_i)\ln\lambda_i,\end{aligned} \tag{25}$$

where

$$P(y_j|x_i)=P(y_j)\exp(sd_{ij})/\lambda_i,\ \lambda_i=\sum_k P(y_k)\exp(sd_{ik}). \tag{26}$$

The $d_{ij}$ is the distortion or loss between $y_j$ and $x_i$; $D$ is the upper limit of the average distortion; $s$=d$R$/d$D$ reflects the slope of $R(D)$. Since s≤0, exp($sd_{ij}$) is between 0 and 1. The exp($sd_{ij}$) can be regarded as truth function $T(\theta_{xi}|y)$, where $\theta_{xi}$ is a fuzzy subset of $V$. Therefore, Equation (26) can be considered as a sematic Bayes' formula (see Equation (21)).

Since $P(y)$ and $P(y|x)$ are interdependent, we need to use an iterative method to produce $P(y)$ [44] (p.326). We may first assume $P(y)$ is flat to obtain $P(y|x)$ from Equation (26) and then use

$$P^{+1}(y_j)=\sum_i P(x_i)P(y_j|x_i) \tag{27}$$

to obtain new $P(y)$, e.g., $P^{+1}(y_j)$. We can repeat Equations (26) and (27) until $P^{+1}(y_j)=P(y)$. This iterative method inspired the author to use it to improve the EM algorithm.

We replace $d_{ij}$ with $I_{ij}$= $I(x_i; y_j)$ =log[$P(x_i|\theta_j)/P(x_i)$] =log[$T(\theta_j|x_i)/T(\theta_j)$] and let $G$ be the lower limit of $I(X; Y_\theta)$. The information rate for given $G$ and source $P(X)$ is defined as

$$R(G) = \min_{P(Y|X): I(X;Y_\theta) \geq G} I(X;Y). \tag{28}$$

Following the derivation of $R(D)$ [41] (p. 31), we obtain

$$\begin{aligned} G(s) &= \sum_i \sum_j I_{ij} P(x_i) P(y_j) 2^{sI_{ij}}) / \lambda_i = \sum_i \sum_j P(x_i) P(y_j|x_i) I_{ij}, \\ R(s) &= sG(s) - \sum_i P(x_i) \log \lambda_i, \end{aligned} \tag{29}$$

where

$$P(y_j|x_i) = P(y_j) P(x_i|\theta_j)^s / \lambda_i,\ \lambda_i = \sum_j P(y_j) P(x_i|\theta_j)^s. \tag{30}$$

Since $I_{ij}$ reflects the verisimilitude of $y_j$ representing $x_i$, we call $R(G)$ the rate-verisimilitude function [36]. Figure 6 shows the comparison of functions $R(D)$ and $R(G)$. Figure 6 (a) shows that $R(D)$ changes with $D$. $R_{max}(D_{min})$=$R(D$=0)=1 bit means that we need at least 1-bit information without distortion. If we allow $D$=$a$/2, then $R(D$=$a$/2)=0, which means we can randomly select $y$ without any information. The classical rate-distortion theory does not consider greater distortion than $D(R$=0)=$a$/2. We can add the dashed line for $D$>$a$/2, which means that if we want to cause enemies' greater loss, we also need certain objective information $R$. For example, $R_{max}(D_{max})$= $R(D$=$a$) =1 means that we need at least 1-bit information to cause loss $a$.

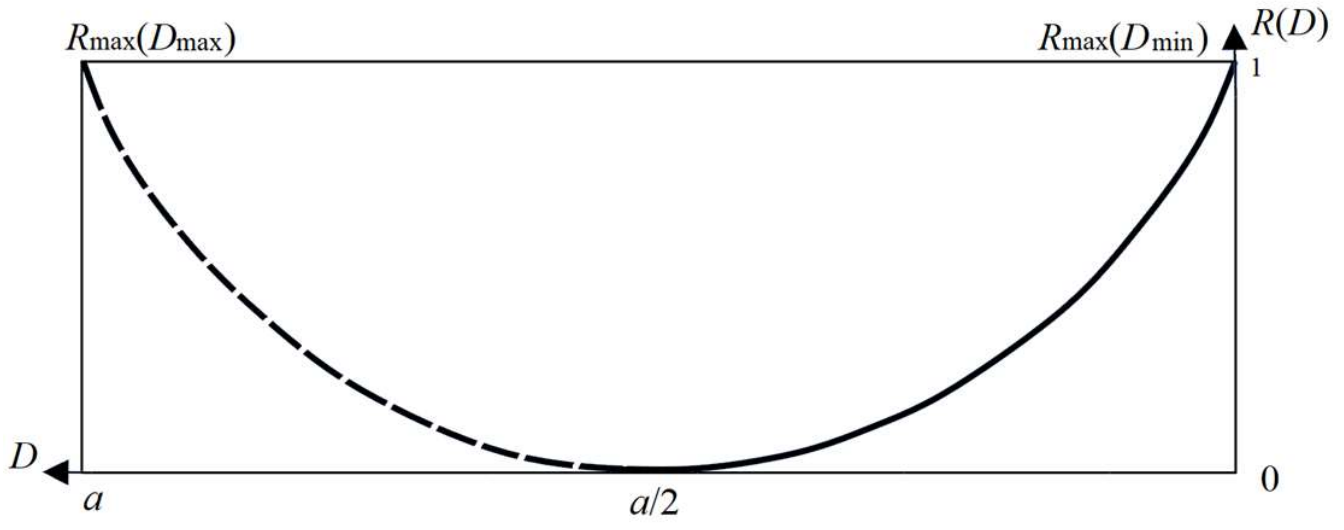

(a) Function $R(D)$ with the extended part (dashed line).

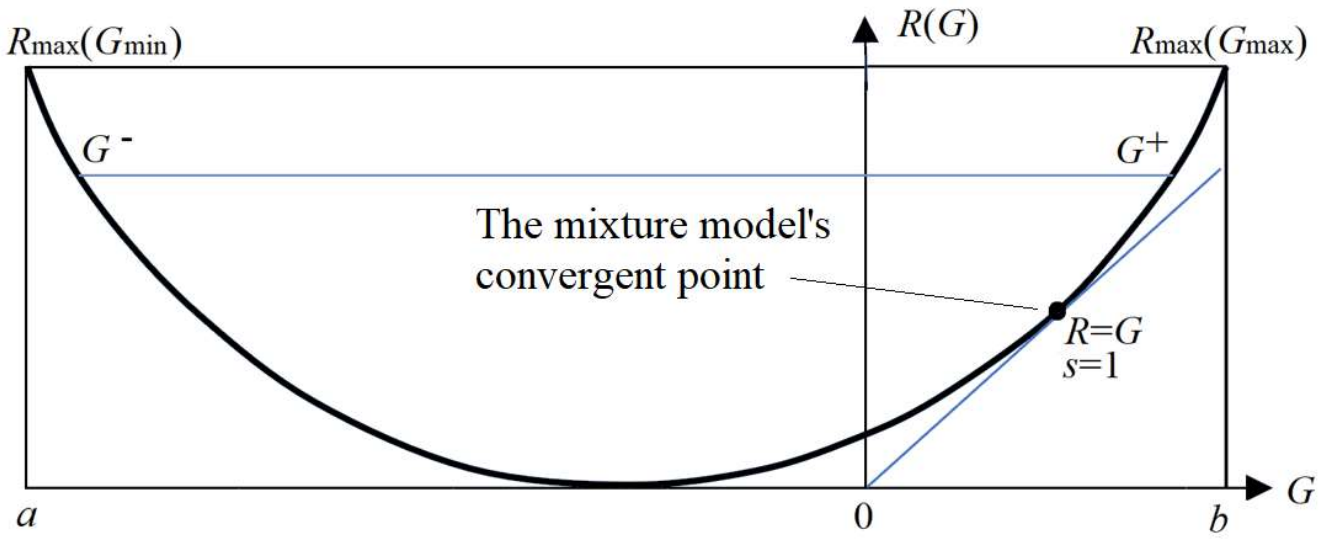

(b)

(c) Function $R(G)$ is used to prove the convergence of the CM-EM algorithm.

Figure 6. Comparing functions $R(D)$ and $R(G)$. $R(G)$ has a matching point $R(G)$=$R$; this point is the convergent point of the mixture model, where $P(x|\theta_j)$=$P(x|y_j)$ for all $j$.

Figure 6 (b) illustrates function $R(G)$. $G^+$ and $G^-$ mean two inverse functions, $G^+(R)$ and $G^-(R)$. Every $R(G)$ function is bowl-shaped with a matching point where $R$=$G$ and information efficiency $G/R$=1. The CM-EM algorithm is to let the semantic channel and the Shannon channel mutually match to achieve $R$=$G$. $R$-$G$ used as the objective function is similar to Jensen–Shannon divergence used as the objective function for Generative Adversarial Network (GAN) [46].

The author used Helmholtz's free energy formula ($F=E-TS$) to explain the information value [33]. The semantic information is similar to the free energy; $s$ is similar to $1/(kT)$ in thermodynamics. The $s$ is also similar to the annealing parameter $\beta$ in the DAEM algorithm. A similar formula in the DAEM algorithm is:

$$P(y_j \mid x) = [P(x \mid \theta_j)P(y_j)]^\beta / \sum_k [P(x \mid \theta_k)P(y_k)]^\beta, j = 1, 2, ..., n. \quad (31)$$

where $\beta$ changes between 0 and 1. However, $s$ changes between $-\infty$ and $\infty$; $s>1$ makes $P(y_j|x)$ steeper and $R$ greater.

In [33] and [37], the author proves that the maximum entropy principle is equivalent to the minimum mutual information principle. He explains information value as free energy, whereas The DAEM algorithm's authors explain the cross-entropy $H_\theta(X)$ as free energy [10,32].

*4.2 The Channel Matching EM (CM-EM) Algorithm*

The relationships between $G$ and other several cross-entropies are:

$$\begin{aligned} &G = I(X;Y_\theta) = H(X) - H(X \mid Y_\theta) = H(X) + H_\theta(Y) - H(Y, X|\theta), \\ &H(X \mid Y_\theta) = -\sum_j \sum_i P(x_i, y_j) \log P(x_i|\theta_j), \\ &H_\theta(Y) = -\sum_i P^{+1}(y_j) \log P(y_j), \end{aligned} \quad (32)$$

where $H(X)$ is the Shannon entropy of $X$, and $H(X|Y_\theta)$ is the posterior cross-entropy of $X$, which is similar to $F$ in the MM algorithm [17]. $H_\theta(Y)$ is the cross-entropy instead of the Shannon entropy. $H(X, Y|\theta)$ is equal to $-Q$. $P^{+1}(y_j)$ comes from Equation (33). The CM-EM algorithm has three steps.

**E1-step:** This step is the same as the E-step of the EM algorithm.

**E2-step:** Improve $P(Y|X)$ by modifying $P(y)$ so that $P^{+1}(y)=P(y)$. This step is to speed mixture proportions' judgments so that the Shannon channel matches the semantic channel. It is similar to the M1-step in the EM algorithm. But it repeats the M1-step several or many times.

We can use the following three equations repeatedly until $P^{+1}(y) = P(y)$.

$$\begin{aligned} &P^{+1}(y_j) = \sum_i P(x_i)P(y_j \mid x_i), j = 1, 2, ..., n; \\ &P(y_j) = P^{+1}(y_j), j = 1, 2, ..., n; \\ &P(y_j \mid x_i) = P(x_i \mid \theta_j)P(y_j) / \sum_k P(y_k)P(x_i \mid \theta_k),\ i = 1, 2, ..., m; j = 1, 2, ..., n. \end{aligned} \quad (33)$$

Repeating Equation (33) makes relative entropy $H(P||P_\theta)$ smaller. In practice, we may repeat Equation (33) no more than $t$ times. The experiments indicate that $t$=3 is good enough in most cases. We call the CM-EM algorithm with $t$=3 as the E3M algorithm. If $t$=1, the CM-EM algorithm becomes the EM algorithm.

Check $H(P||P_\theta)$. If $H(P||P_\theta)$ is less than a tiny value, such as 0.001 bit for a continuous sampling distribution, or 0.01 bit or the like for a sampling distribution from a sample of a specific size, then end the iteration. We can also stop when some parameters' increments are smaller than some tiny values.

**MG-step** (for maximizing $G$): This step is the same as the M2-step in the EM algorithm. With Equation (24), we fix $P(y_j|x)$ and optimize the model parameters in $P(x|\theta_j)$ to maximize $G$. Then, go to E1-step. $G$ reaches the maximum when $P(x|\theta_j)=P(x|y_j)$ for all $j$. Hence, the new likelihood function is

$$P(x|\theta_j^{+1}) = P(x)P(x|\theta_j)/P_\theta(x). \quad (34)$$

This step lets the semantic channel match the Shannon channel so that $T(\theta_j|x) \propto P(y_j|x)$ for all $j$. If $P(x|\theta_j)$ is a Gaussian distribution, we can obtain new parameters:

$$\mu_j^{+1}=\sum_i P(x_i\,|\,\theta_j^{+1})x_i,\ j=1,2,...,n;$$
$$\sigma_j^{+1}=\{\sum_i P(x_i\,|\,\theta_j^{+1})[x_i-\mu_j^{+1}]^2\}^{0.5},\ j=1,2,...,n. \tag{35}$$

If the likelihood functions are not Gaussian functions, we can find the optimized parameters by searching the parameter space or using the Gradient Descent.

For a sample with size $N$: $\{x(t)\,|\,t=1,2,\ldots N,\ x(t)\in U\}$, we may also use

$$\mu_j^{+1}=\frac{1}{N}\sum_{t=1}^{N}P(x(t)\,|\,\theta_j^{+1})x(t),\ j=1,2,...,n;$$
$$\sigma_j^{+1}=\{\frac{1}{N}\sum_{t=1}^{N}P(x(t)\,|\,\theta_j^{+1})[x(t)-\mu_j^{+1}]^2\}^{0.5},\ j=1,2,...,n. \tag{36}$$

If $U$ is two-dimensional, we need to calculate the correlation coefficients.

*4.3. The Convergence Proof for the EM and CM-EM Algorithms*

This proof makes use of the properties of function $R(G)$ as shown in Figure 6 (b):

- $R(G)$ is concave, and $R(G)$-$G$ has the exclusive minimum 0 as $R(G)=G$;
- $R(G)$-$G$ is close to relative entropy $H(P||P_\theta)$.

After the E1-step (or the E-step in the EM algorithm), Shannon's mutual information $I(X;\ Y)$ becomes

$$R=\sum_i\sum_j P(x_i)\frac{P(x_i\,|\,\theta_j)}{P_\theta(x_i)}P(y_j)\log\frac{P(y_j\,|\,x_i)}{P^{+1}(y_j)}. \tag{37}$$

We define

$$R''=\sum_i\sum_j P(x_i)\frac{P(x_i\,|\,\theta_j)}{P_\theta(x_i)}P(y_j)\log\frac{P(x_i\,|\,\theta_j)}{P_\theta(x_i)}. \tag{38}$$

It is easy to prove $R''$-$G$=$H(P||P_\theta)$. Hence

$$R=\sum_i\sum_j P(x_i)\frac{P(x_i\,|\,\theta_j)}{P_\theta(x_i)}P(y_j)\log\left[\frac{P(x_i\,|\,\theta_j)P(y_j)}{P_\theta(x_i)P^{+1}(y_j)}\right]=R''-H(P_y^{+1}||P_y),$$
$$H(P_y^{+1}||P_y)=\sum_j P^{+1}(y_j)\log[P^{+1}(y_j)/P(y_j)]; \tag{39}$$

$$H(P\,||\,P_\theta)=R''-G=R+H(P_y^{+1}||P_y)-G. \tag{40}$$

Equation (40) is the basic formula for the convergence proof. The three steps in the CM-EM seemly just improve $R$, $H(P_y^{+1}||P_y)$, and $G$, respectively. However, the convergence proof's difficulty is that when we minimize $R$ or $H(P_y^{+1}||P_y)$, the other two items also change. For example, when we minimize $H(Y^{+1}||Y)$, we do not know whether it is possible that $R$-$G$ increases too much to reduce $R''$-$G$.

**The Convergence Proof:** Proving that $P_\theta(x)$ converges to $P(x)$ is equivalent to proving that $H(P||P_\theta)$ converges to 0. Since E2-step makes $R$=$R''$ and $H(P_y^{+1}||P_y)$=0, we only need to prove that every step minimizes $R$-$G$ after the beginning step (the first E1 step).

The MG-step minimizes $R$-$G$ because this step maximizes $G$ without changing $R$. The remaining task is to prove that E1-step and E2-step minimize $R$-$G$. Fortunately, we can strictly prove that by variational and iterative methods that Shannon [42] and others [43,44] used for analyzing the rate-distortion function. The following analysis is a little different from the rate-distortion function analysis. We replace distortion $d_{ij}$ with predictive information $I(x_i;\ \theta_j)$=log[$P(x_i\,|\,\theta_j)/P(x_i)$].

We use the Lagrangian multiplier method to optimize $P(y|x)$ and $P(y)$ to minimize $R$-$G$=$I(X; Y)$ - $I(X; Y_\theta)$. Since $P(y|x)$ and $P(y)$ are interdependent, we can only fix one to optimize another. To optimize $P(y|x)$ and $P(y_j)$, two restrictive conditions are

$$\sum_j P(y_j \mid x_i) = 1,\ i = 1, 2, ..., n;\ \ \sum_j P(y_j) = 1. \tag{41}$$

The Lagrange function is therefore

$$F = I(X;Y) - I(X;Y_\theta) + \sum_i \mu_i \sum_j P(y_j \mid x_i) + \alpha \sum_j P(y_j). \tag{42}$$

To optimize $P(y|x)$, we fix $P(y_j)$ in $F$ and order $\partial F / \partial P(y_j \mid x_i) = 0$. Then we derive the optimized $P(y|x)$ (see Appendix D for details):

$$P^*(y_j \mid x_i) = P(y_j) P(x_i \mid \theta_j) / \sum_k P(y_k) P(x_i \mid \theta_k),\ i=1, 2, ..., n;\ \ j=1, 2. \tag{43}$$

This formula is used in the EM algorithm's E-step and the CM-EM algorithm's E1-step. Thus, the E-step and the E1-step minimize $R$-$G$. To optimize $P(y)$, we fix $P(y_j|x_i)$ in $F$ and then order $\partial F / \partial P(y_j) = 0$. Hence, we derive the optimized $P(y)$ (see Appendix D for details):

$$P^*(y_j) = \sum_i P(x_j) P(y_j \mid x_i),\ j=1, 2, ..., n. \tag{44}$$

This formula is used in the EM algorithm's M-step and the CM-EM algorithm's E2-step. Thus, every M-step and every E2-step minimize $R$-$G$. According to Equation (40), and

- every step minimizes $R$-$G$,
- the E2-step reduces $H(P_y^{+1}||P_y)$ to 0, and
- $R(G)$-$G$ is concave with the exclusive minimum 0 as $R(G)$=$G$ (see Figure 6 (b)),

the relative entropy can converge to its minimum. **QED**.

The above convergence proof is not limited to Gaussian mixture models. For other mixture models, the convergence proof is the same. If every MG-step can increase $G$ in every iteration, then $H(P||P_\theta)$ can converge to its global minimum.

Beal [18] used a variational method earlier to derive the same $P^*(y_j|x)$ in proving that the E-step maximizes $L$. This proof is significant. However, since $P(y|x)$ is also the function of $P(y)$, it is not enough to talk about optimizing $P(y|x)$ without mentioning $P(y)$.

### *4.4. Relationships between the EM, MM, and CM-EM Algorithms*

In the EM algorithm's M-step, maximizing $Q$ is equivalent to minimizing the cross-entropies $H_\theta(Y)$ and $H(X|Y_\theta)$ because

$$Q = -H(X, Y \mid \theta) = -H_\theta(Y) - H(X \mid Y_\theta). \tag{45}$$

Since $G$=$H(X)$-$H(X|Y_\theta)$, minimizing $H(X|Y_\theta)$= is equivalent to maximizing $G$. Therefore, there is the relationship:

The E-step of the EM algorithm = The E1-step of the CM-EM algorithm,

The M1-step of the EM algorithm ≈The E2-step of the CM-EM algorithm,

The M2-step of the EM algorithm=The MG-step of the CM-EM algorithm.

The difference is that the M1-step only modifies $P(y)$ one time, whereas the E2-step modifies $P(y)$ many times until $P^{+1}(y)$=$P(y)$ or several times until $P^{+1}(y)$≈$P(y)$ so that $P(y|x)$ is a proper Shannon channel.

Since the CM-EM algorithm becomes the EM algorithm when the E2-step only runs Equation (33) one time ($t$=1), the convergence proofs for both algorithms are almost the same.

Consider the MM algorithm provided by Neal and Hinton [17]. The objective function is

$$F = Q + H(Y) = -H(X, Y \mid \theta) + H(Y) \approx -H(X \mid Y_\theta). \tag{46}$$

The "≈" is used because of $P^{+1}(y) \neq P(y)$ and hence $H_\theta(Y) \neq H(Y)$ in general. If we replace $H(Y)$ with $H_\theta(Y)$ for $F$, then $F$ is equal to $-H(X|Y_\theta)$. Since $G=H(X)-H(X|Y_\theta)$, maximizing $F$ in the MM algorithm's M-step is similar to maximizing $G$ in the MG-step. However, the E-step of the MM algorithm also maximizes $F$. The E1 and E2-steps of the CM-EM algorithm are different. The reason is that $G$ might be bigger than $G^*$, and hence, $G$ and $F$ should be decreased (see Figure 2 (c) and Figure 5).

According to Equation (45), if $H_\theta(Y)$ increases with $L$, $Q$ will decrease. The experiments show that $H_\theta(Y)$ may increase with $L$, such as in a case where $P^*(y_1)= P^*(y_2)=0.5$, whereas $P(y_1)=0.9$ and $P(y_2)=0.1$.

## 5. Results: More Experiments with Both Algorithms and the Fair Competition Principle

### *5.1. More Experiments about Example 2*

Using sample $\{x(1), x(2),\ldots,x(N)\}$ (e.g., using Equation (36)) to update parameters or using sampling distribution $P(x)$ (e.g., using Equation (35)) to optimize parameters, the iterative speeds are different. When the sample is small, the former (Equation (36)) is good. But when the sample is big, such as $N$=50000, the former might need ten times of time that the latter (Equation (35)) needs. In the following experiments, the sample distribution $P(x)$ was used.

The author has used different initial means to compare the iteration numbers of the EM and E3M (CM-EM with $t$=3) algorithms for Example 2 (see Figure 7). The true model is $(\mu_1^*, \mu_2^*, \sigma_1^*, \sigma_2^*, P^*(y_1))$ =(100, 125, 10, 10, 0.7). The initialization is $(\mu_1, \mu_2, \sigma_1, \sigma_2, P(y_1))$=( $\mu_1, \mu_2$, 7, 7, 0.5). The sample size $N$ is 50000. Two numbers "289/191" at the lower-left corner ($(\mu_1, \mu_2)$=(80, 81)) means that using an initial point $(\mu_1, \mu_2)$ there, the EM algorithm needs 289 iterations, and the E3M algorithm needs 191 iterations. The other numbers are in like manner. For every initial point, each algorithm was repeated three times; the middle iteration number was selected. The stop condition was that $|\max(\mu_1, \mu_2)-\max(\mu_1^*, \mu_2^*)|<1$, $|\min(\mu_1, \mu_2)-\min(\mu_1^*, \mu_2^*)|<1$, $|\sigma_1-\sigma_1^*|<1$, $|\sigma_2-\sigma_2^*|)<1$, $|\max(P(y_1), P(y_2))-\max(P^*(y_1), P^*(y_2))|<0.033$, and $H(P||P_\theta)<0.005$. Note that in practices, we may only use $H(P||P_\theta)<0.005$ as the stop condition without considering the symmetry. We can also put iteration numbers on the lower-right side according to the axisymmetry. For the iteration numbers at the absolute equality line, $(\mu_1, \mu_2)$= $(\mu_1, \mu_1+1)$ was used as the initial means.

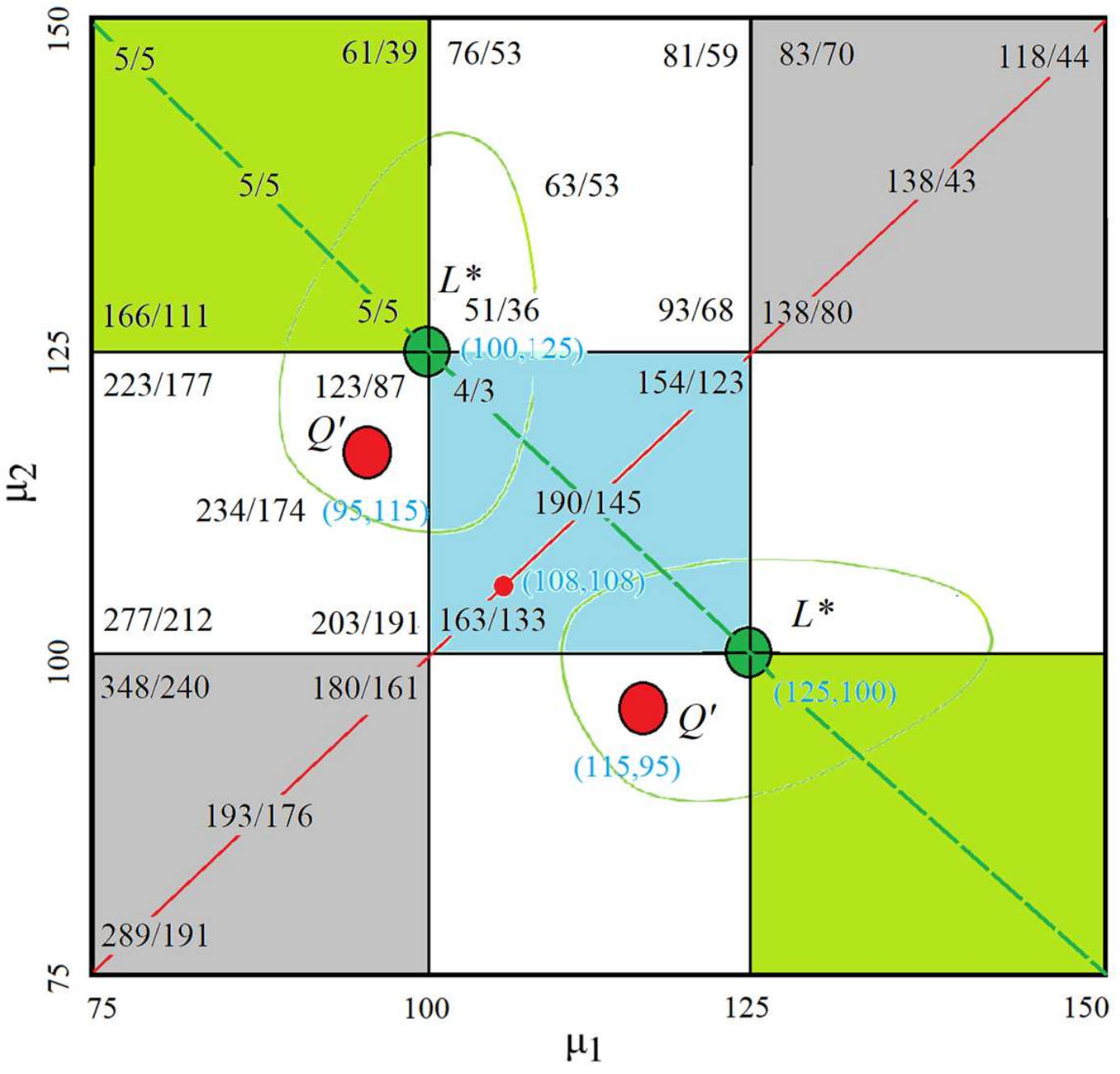

Figure 7. Iteration numbers change with different initial means for Example 2. "289/191" means that EM's iteration number is 289, and E3M's is 191. The sample size is 50000.

The average iteration numbers are 136.7 (with EM algorithm) and 90.4 (with E3M algorithm). The E3M algorithm needs 74% of the iterations that the EM algorithm needs. When sample sizes are not big enough, the convergence points are very uncertain.

The author used Example 2 with different sample sizes and different initial means to examine the EM and E3M algorithms. For every initial condition, two algorithms run 20 times. Table 1 shows the results. The stop condition for $N$=1000 is that $|\mu_i-\mu_i^*|<1$ and $|\sigma_i-\sigma_i^*|<1$, $j$=1,2. The stop condition for $N$=100 is that $|\mu_i-\mu_i^*|<3$ and $|\sigma_i-\sigma_i^*|<3$, $|P(y_j)-P^*(y_j)|<0.1$, $j$=1,2; $H(P||P_\theta)<0.7$ bit. The samples are produced by a random function so that any two samples in different runs are different.

**Table 1**. Comparison between the EM and CM-EM algorithms

| | $N$ | Initial ($\mu_1$, $\mu_2$) | Initial $\sigma_1$=$\sigma_2$ | Times of invalid convergence | Times of IN>500 | Times of fast convergence |
|---|---|---|---|---|---|---|
| EM | 1000 | (95,115) unfair | 10 | 2 | 7 | 5 (IN<100) |
| EM | 1000 | (95,115) | 20 | 1 | 9 | 5 (IN<100) |
| EM | 1000 | (95,115) | 5 | 1 | 9 | 6 (IN<100) |
| CM-EM | 1000 | (95,115) | 10 | 1 | 5 | 10 (IN<100) |
| CM-EM | 1000 | (95,115) | 5 | 0 | 4 | 12 (IN<100) |
| EM | 100 | (80,145) fair | 10 | 3 | 3 | 16 (IN<4) |
| CM-EM | 100 | (80,145) | 10 | 1 | 1 | 19 (IN<4) |
| CM-EM | 100 | (80,145) | 5 | 2 | 2 | 18 (IN<4) |

* IN means an iteration number; $N$ is the sample size.

When the sample size is 1000, the EM algorithm performs similarly as $\sigma$=5 and as $\sigma$=20. When the sample size becomes 100 and the initial means are fair, both algorithms have fewer invalid convergence times and faster convergent speeds than before. Under every condition, the CM-EM algorithm outperforms the EM algorithm.

The author used the CM-EM algorithm and the FCP for Example 2, as shown in Figure 5. After the E3M algorithm's first iteration, one mixture proportion became a tiny value (0.009). Then he changed this blocked component's means to another side of the sample (e.g., change GGLL to GLLG) to speed the convergence. The iteration number was less than 10.

### *5.2. Using Two Open Datasets to Compare the E3M and EM Algorithms*

For researchers to repeat the same experiment, the author used two open datasets named g2-1-50 and g2-2-50 ([44], g2-1-50.txt and g2-2-50.txt in G2 sets) to compare the two algorithms.

***Example 3.*** *The dataset g2-1-50 is used as the observed data. The sample size is 2048. The true model that produced g2-2-50 is* ($\mu_1^*$, $\mu_2^*$, $\sigma_1^*$, $\sigma_2^*$, $P^*(y_1)$)=(500, 600, 50, 50, 0.5).

First the author used the initial model ($\mu_1$, $\mu_2$, $\sigma_1$, $\sigma_2$, $P(y_1)$)=(450, 550,50, 50, 0.3). Figure 8 shows two initial components in Figure 8 (a) and two convergent components in Figure 8 (b). Table 5 provides more comparisons.

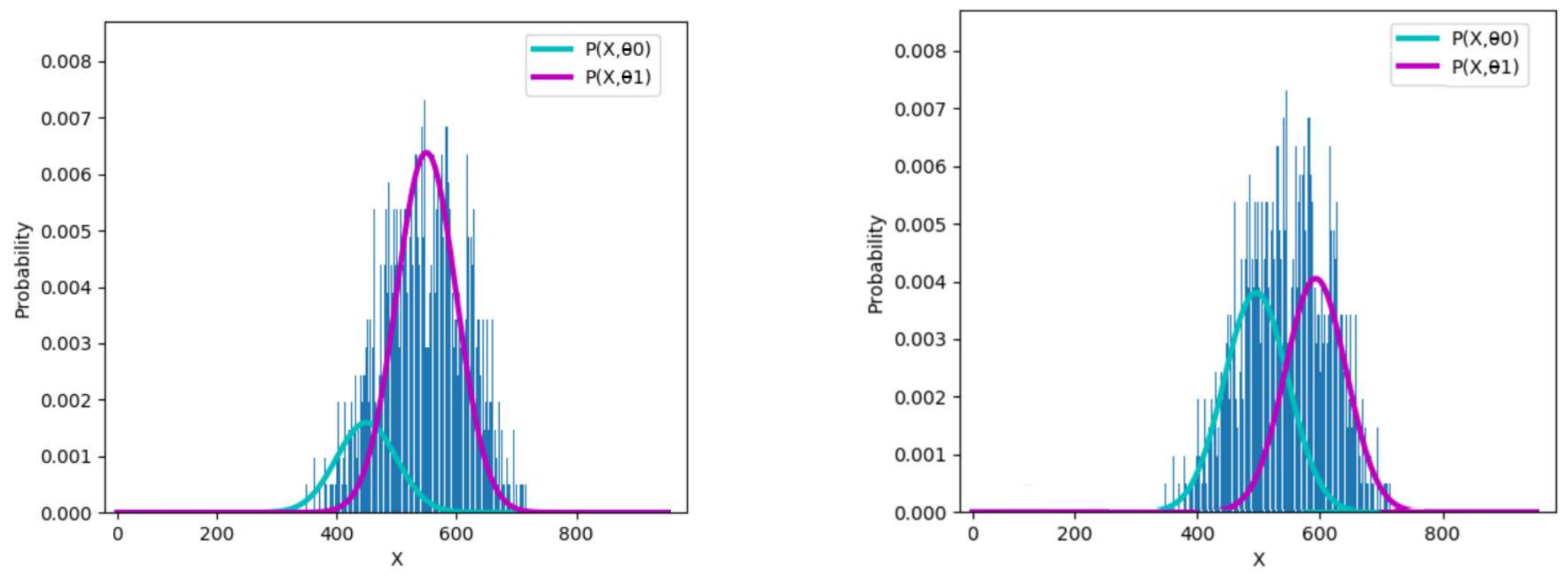

(a) initial components (b) convergent components

Figure 8. Using dataset g2-1-50 to compare the E3M and the EM algorithms. The EM algorithm needs 274 iterations, whereas the E3M algorithm needs 178 iterations for relative entropy<0.138 bits.

**Table 2.** Using dataset g2-1-50 to compare the EM and E3M algorithms

| Initialization ($\mu_1$, $\mu_2$, $\sigma_1$=$\sigma_2$, $P(y_1)$) | Iteration Number with EM | Iteration Number with E3M | Competition |
|---|---|---|---|
| (450, 550, 50, 50, 0.3) | 253 | 178 | Unfair |
| (450, 55, 50, 50, 0.2) | 274 | 181 | Unfair |
| (450, 650, 50, 50, 0.5) | 5 | 5 | Fair |
| (450, 600, 50, 50, 0.5) | 136 | 106 | Unfair |

The results in the last row are unexpected. The last model (450, 600, 50, 50, 0.5) is closer to the true model than the third model (450, 650, 50, 50, 0.5); however, the former is much worse than the latter. These results reveal that fair competition between components is more critical than coincidences between components and sub-samples. There is a similar situation in Example 5.

For the initial model in the first line, the E5M algorithm needs 167 iterations; nevertheless, the E3M algorithm is more economical than the E5M algorithm.

***Example 4****. The dataset g2-2-50 [44] is used as the observed data. The sample size is 2048. The true model that produced g2-2-50 is* ($\mu_{1m}$*, $\mu_{1n}$*, $\mu_{2m}$*, $\mu_{2n}$*, $\sigma_{1m}$*, $\sigma_{1n}$*, $\sigma_{2m}$*, $\sigma_{2n}$*, $r_1$*, $r_2$*, $P^*(y_1)$)=*(500, 500, 600, 600, 50, 50, 50, 50, 0, 0, 0.5), where m and n are two coordinate names;* $r_1$ *and* $r_2$ *are correlation coefficients.* The initial model is ($\mu_{1m}$, $\mu_{1n}$, $\mu_{2m}$, $\mu_{2n}$, $\sigma_{1m}$, $\sigma_{1n}$, $\sigma_{2m}$, $\sigma_{2n}$, $r_1$, $r_2$, $P(y_1)$)=(500, 500, 700, 700, 22, 22, 22, 22, 0, 0, 0.5). *The stop condition is the deviation of every location parameter (with* $\mu$*) is less than or equal to 10.*

Figure 8 shows the results.

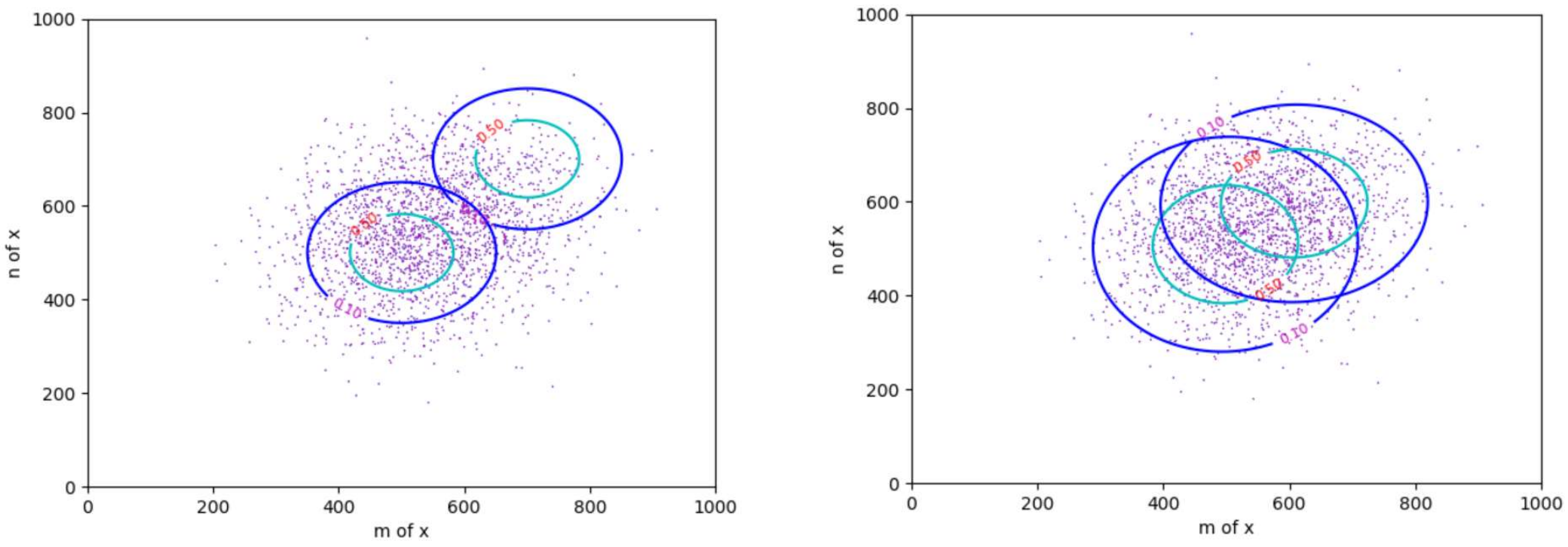

(a) Initial components (b) Global convergence

**Figure 9**. Using dataset g2-2-50 to compare the E3M and EM algorithms. The EM algorithm needs 144 iterations, whereas the E3M algorithm needs 98 iterations so that each $\mu$'s deviation is less than 10.

If we use the FCP to change ($\mu_{1m}$, $\mu_{1n}$)= (500, 500) into ($\mu_{1m}$, $\mu_{1n}$)= (400, 400), both the EM and E3M algorithms can converge faster than before.

*5.3. Comparing the Running Times of the EM and E3M Algorithms*

For comparing the EM and E3M algorithms' running times, the author used Example 2 with the sample size $N$=50000. The initialization was ($\mu_1$, $\mu_2$, $\sigma_1$, $\sigma_2$, $P(y_1)$)=(80, 130, 10, 10, 0.5). Two typical results are shown in Table 3.

**Table 3.** Comparing running times (seconds) of the EM algorithm and the E3M algorithm

| | Iteration number | Total time (seconds) | Time per iteration (seconds) |
|---|---|---|---|
| EM algorithm | 126 | 12.9 | 0.103 |
| E3M algorithm | 89 | 9.92 | 0.111 |

Since the E3M algorithm needs about 76% of the iterations that the EM algorithm needs, the E3M algorithm needs about 0.76*1.08=82% of the running time that the EM algorithm needs on average. Therefore, the E3M algorithm can save about 18% of the running time that the EM algorithm needs.

Using sample {$x(1)$, $x(2)$,…,$x(N)$} (e.g., using Equation (36)) to update parameters or using sampling distribution $P(x)$ (e.g., using Equation (35)) to optimize parameters, the iterative speeds are different. When the sample is small, the former (Equation (36)) is good. But when the sample is big, such as $N$=50000, the former might need ten times the latter (Equation (35)) needs.

### *5.4. A Two-Dimensional Example for Explaining the Influences of Sample sizes and Fair Competition*

Example 5 was used to examine the influences of samples' sizes and fair competition for a two-dimensional mixture model. Both the EM and CM-EM algorithms were used.

***Example 5****. Three pairs of samples were used (see Figure* ***10****). The sample size was 1000. The stop condition is* $|\mu_{mj}-\mu_{mj}^*|<1$ and $|\sigma_{mj}-\sigma_{mj}^*|<1$, $j$=1, 2, ⋯, 6. *The True and initial parameters for Figures 10-12 can be found in Supplemental Materials.*

Figure 10 (a) shows an unfair initialization so that the upper pair and the lower pair of components could not validly converge. After the sample size became 10000 (see Figure 10 (c)) or the initial means were adjusted (see Figure 10 (b)), all components could validly converge (see Figure 10 (c) and (d)).

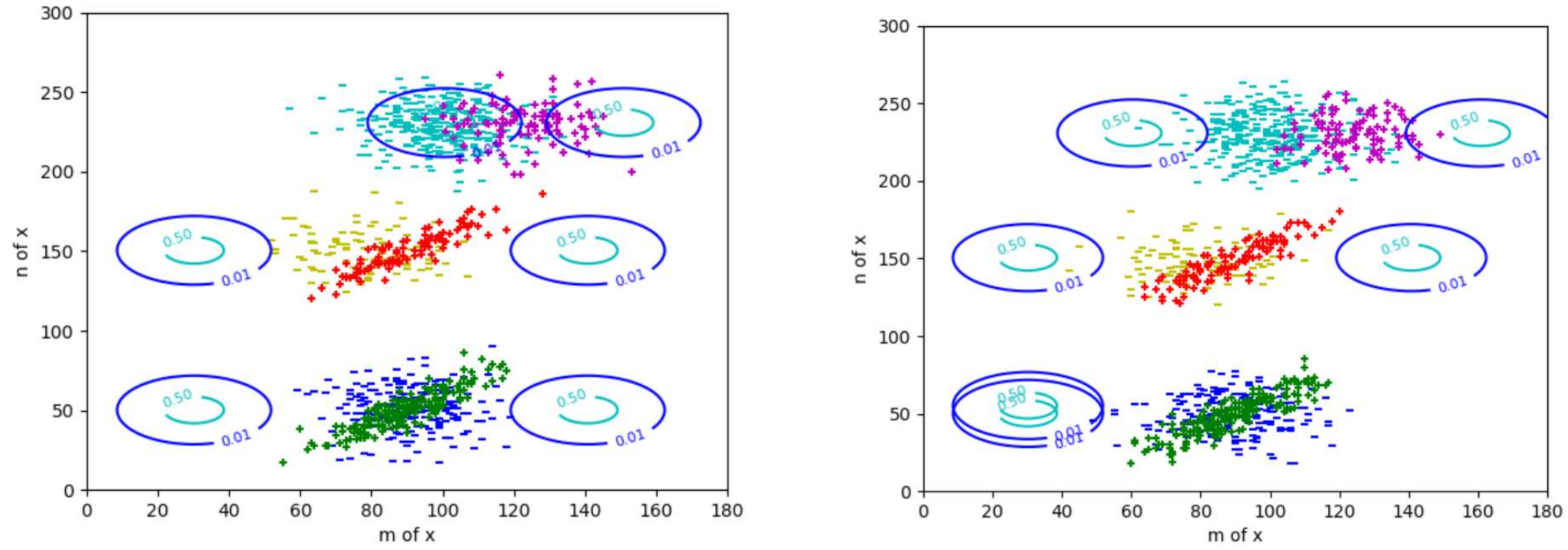

(a) Unfair initial means. (b) Fair initial means.

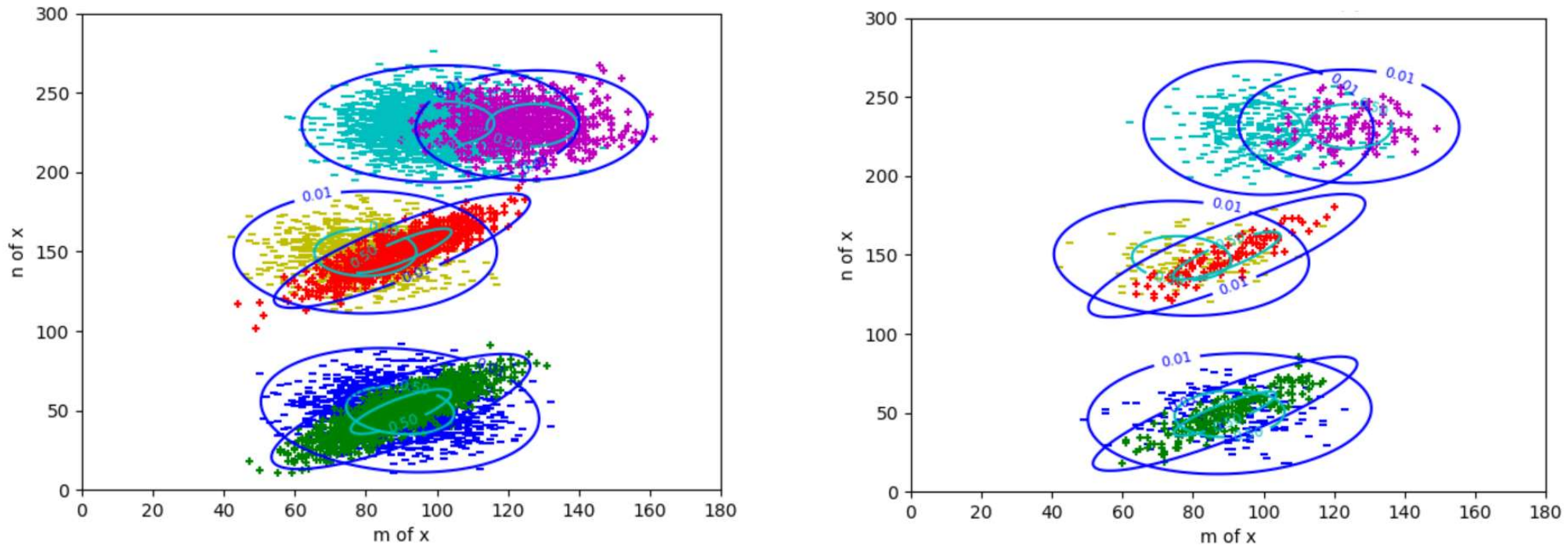

( c) Valid convergence as the sample size was 10000. (d) Valid convergence for fair initial means.

Figure 10. The influences of sample sizes and fair competition. For the convergence in (c), the E3M algorithm spent 50-60 iterations, whereas the EM algorithm spent 70-100 iterations.

This example indicates that even if the overlap is severe, the EM algorithm can also validly converge if the sample size is big enough or the initial means are fair. However, for some improperly initial parameters and smaller samples, the convergence might be invalid. This example also indicates that the E3M algorithm may need fewer iterations than the EM algorithm for unfair initial means.

This example also explains that fair competition between components is more critical than the coincidence between components and sub-samples.

### 5.6. A Two-dimensional Example for Avoiding Boundary Convergence

The following sample was obtained from an open dataset named s3 ([47], s3.txt in S-sets). We use this example to explain that the FCP and the CM-EM algorithm's E2-step are helpful for the EM algorithm to step out of the boundary convergence.

***Example 6.*** *In a two-dimensional sample space, some components are blocked by others, as shown in Figure 11. The task is to find and move the blocked components for global convergence.*

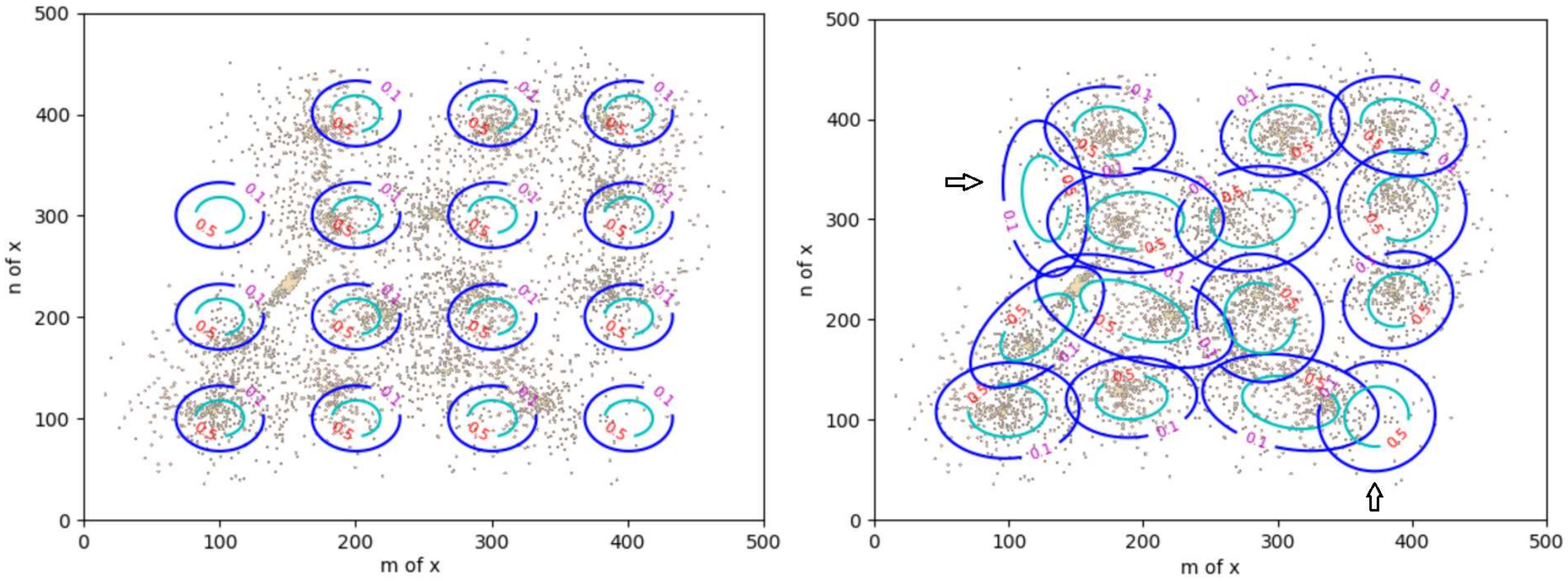

(a) Set initial means simply.

(b) Finding blocked components after the first E2-step.

**Figure 11**. Illustrating how to find some blocked components by the first iteration of the CM-EM algorithm. After the first iteration, the mixture proportions of two blocked components pointed by two arrows become small.

The author first set initial means simply, as shown in Figure 11 (a). After one iteration with the E3M algorithm, the mixture proportions of two blocked components decreased to two tiny values (0.0071 and 0.0122). Then, it was easy to find blocked components.

In Figure 12 (a), the three components were moved to proper positions, according to the FCP, so that all components can globally converge, as shown in Figure 12 (b).

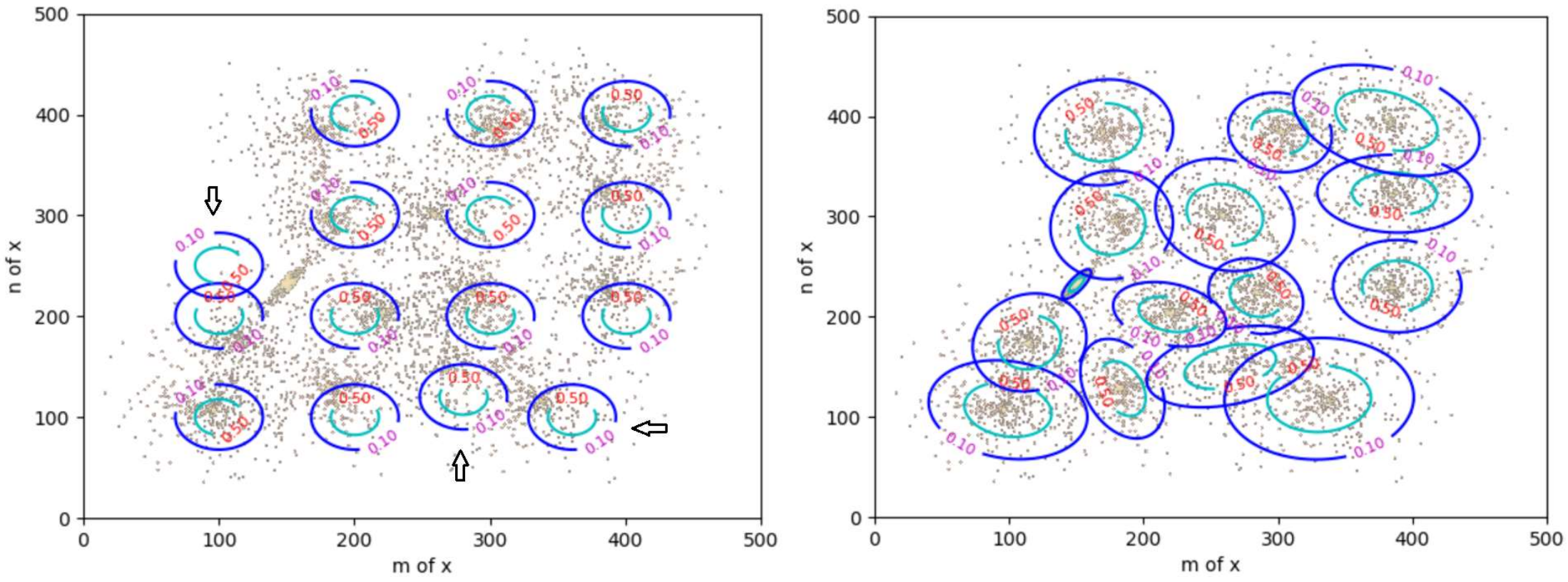

(a) Starting with fair means.

(b) Global convergence after 30 iterations.

**Figure 12**. Illustrating how to move the blocked components (a) for the global convergence (b). Three arrows point to the three moving components in (a). Both the EM and E3M algorithms need 25 iterations so that relative entropy reaches 4.302 bits.

In this example, fair initial means are most important; the two algorithms perform similarly. We shall discuss boundary convergence again in Section 6.3.

## 6. Discussions

### *6.1. Theoretical Significance of This Study*

This study provides the FCP and the IEM theory for mixture models. The FCP explains why different initial parameters cause various convergence difficulties. The initialization map based on this principle provides a new method for mixture models' initializations.

The core thought in the *Q-L* synchronization theory basing the DAEM algorithm is to maximize *Q* repeatedly so that the log-likelihood *L* reaches its maximum. In contrast, the IEM theory's core thought is to maximize information efficiency *G*/*R* or minimize *R*-*G* repeatedly so that $H(P||P_\theta)$ approaches 0 or the minimum. This paper proposed and tested a new convergence theory of mixture models. This theory has been strictly proved by the above mathematical analysis in Section 4.3 and tested by the six examples in Section 3 and Section 5.

Without a correct convergence theory, an improved EM algorithm may perform better for some examples and worse for other unknown examples. For example, all improvements for the global maximum of *Q* or *F* are improper because, in some cases, *Q* or *F* should decrease (see Figure 2 (c) and Figure 5 (c)). The new convergence theory reminds us to avoid blind improvements to the EM algorithm.

### *6.2. About How Results Support Hypotheses*

The author claims Affirmation I and Affirmation II (see Section 1) in the *Q-L* synchronization theory are wrong; we can improve the EM algorithm using the FCP and the CM-EM algorithm.

Example 1 and Example 2 in Section 3 reveal Affirmation I is wrong because the observed data log-likelihood *L* and the complete data log-likelihood *Q* are not always positively correlated. And Affirmation II is false because *Q* may and should decrease in some cases. Section 3.1 provides the mathematical analysis that explains why the global maximum of *Q* is not $Q^*$ that results in the globally maximal *L*. Sections 3.3 and 3.4 indicate that local convergence happens not because of locally maximal *Q* but because of unfair competition and small samples.

The CM-EM algorithm is provided as an improved EM algorithm. Sections 5.1 and 5.2 compare the EM and CM-EM algorithms' convergence difficulties for the typical examples (Examples 2 -4) with various initial means. The results indicate that the CM-EM algorithm can save about 26% of the iterations that the EM algorithm needs. We can accelerate convergence using the FCP and the initialization map. Examples 5 and 6 indicate that the FCP helps the global convergence of two-dimensional Gaussian mixture models. Example 6 demonstrates that the FCP and the CM-EM algorithm's E2-step can help us avoid boundary convergence.

As for running time for the typical example (Example 2), the E3M algorithm can save about 18% of the time that the EM algorithm needs on average (see Section 5.3). If we use the initialization map (Figure 7), both EM and CM-EM algorithms can vastly save iteration numbers and running times.

The initialization map with two groups of iteration numbers in Figure 7 provides a method for the comprehensive comparison between an improved EM algorithm and the EM algorithm. Some improved EM algorithms seem better than the EM or CM-EM algorithms because they can save 30%-60% of the iterations [19, 23, 24] that the EM algorithm needs. However, these comparisons are not sufficient because an improved EM algorithm may be better in some cases and worse in other cases. For example, in Example 1 (see Figure 2 (c)), where *F* and *G* decrease in every iteration, the MM algorithm will be hard to converge because it also maximizes *F* in the E-step. Without a comprehensive comparison, we cannot obtain fair conclusions.

### 6.3. About the Boundary Convergence and the CM-EM Algorithm's Limitation

Regarding how to avoid the local convergence and the boundary convergence, there have been many useful methods, such as the SMEM algorithm [13], the CEM algorithm [9], the Random swap EM algorithm [14], and the Cross-entropy algorithm [15]. In these algorithms, exterior circulations are used to improve initial or minor parameters. In the Random Swap EM algorithm, the method is to remove a component and add a component randomly. In the SMEM algorithm, the way is to split two components and merge two components. In the CEM algorithm, selecting split and merge operations is based on the competitive mechanism.

Using the FCP introduced in Sections 3.3 and 3.4, we can easily avoid the local convergence and accelerate the global convergence of the binary Gaussian mixture. However, for the mixtures of more components or mixtures in two-dimensional or multi-dimensional sample spaces, the FCP only is not sufficient. It is still a challenge to enrich this principle and convert it to practical methods. Nevertheless, the FCP and the CM-EM algorithm's E2-step does not repel other improvements to the EM algorithm. How do we combine the existing algorithms with the FCP and the CM-EM algorithm's E2-step for better convergence? We need further studies.

### 6.4 About the Self-annealing Behavior of the EM Algorithm

Yu et al. [48] researched better annealing parameter $\beta$ for the DAEM algorithm. They prove that the EM algorithm exhibits a self-annealing behavior in some cases mentioned by Figueiredo and Jain earlier. For example, when all means $\mu_1$, $\mu_2$, … are the same, there is local convergence to the sample's mass center. In Figure 4, we show a similar conclusion that an initial point at the 45° line or in the deep ditch must converge to $f$, but ($\mu_1$, $\mu_1$+1) will converge to $a_2$. According to the $Q$-Proof, a local convergent point $f$ in the parameter space is a stable fixed point. However, this self-annealing behavior indicates that the points near the 45° line or point $f$ are not attracted to "the stable fixed point" $f$. Therefore, this self-annealing behavior is not compatible with the $Q$-Proof. Saâdaoui [23] provides a similar analysis that supports global convergence. The author of this paper thinks that the above-mentioned self-annealing behavior is not to avoid locally maximal $Q$ but to get away from $Q$'s deep ditch. The EM algorithm does have the self-annealing behavior for escaping from locally maximal $Q$, such as in Figure 1 for Example 1 and Figure 5 for Example 2. The EM algorithm cannot escape from the local convergence only because the sample is too small, and unfair competition exists.

Yu et al. [45] optimize the annealing parameter $\beta$ to accelerate the DAEM algorithm. Perhaps we may follow their method to optimize $s$ (see Equation (28)) to speed the CM-EM algorithm. When $R<R^*$, $s>1$ should be better; when $R>R^*$, $s<1$ should be better. But the author's experiments indicated that changing initial means was much easier to speed the convergence than changing $s$ or $\beta$.

### 6.5. About Variational EM Algorithms

Most Variational EM (VEM) algorithms also use Bayesian Inference and Bayesian prior $q(\theta)$ [18, 24, 25]. The objective function used by Beal [18] is $L$ or its lower boundary $\boldsymbol{F}(P(y|x), q(\theta))$. Beal's variational method is to optimize $P(y|x)$. The objective functions used by Assad et al. [24] and Liu et al. [25] are $F=Q+H(Y)$ proposed by Neal and Hinton [17]. Their variational method is to optimize $P(y)$. The above VEM algorithms have been well applied to some areas, such as temporal data clustering [24] and non-Gaussian mixture [25].

The CM-EM is also a VEM algorithm. Its objective function is $R$-$G$. We can minimize $H(P||P_\theta)$ or maximize $L$ by minimizing $R$-$G$. This variational method is to optimize both $P(y|x)$ and $P(y)$. Since the E1-step and the E2-step do not change model parameters and do not use Bayesian inference with $P(\theta)$, the calculation is straightforward. In comparison with the above VEM algorithms, the CM-EM algorithm optimizes mixing proportions more quickly. In Figure 2 and Figure 5, since $G$ and $F$ should decrease in some E-steps, it is not suitable to maximize $F$ in the E-step of a VEM algorithm.

### 6.6. About Maximum Mutual Information Classifications

After we obtain the optimized components or several mixture models from a clustering algorithm, we need to provide a classifier $z = f(x)$ with a criterion to classify the sample space $U$, where $z$ is the prediction of $y$. The maximum mutual information criterion is becoming famous for feature space classification. The Channel Matching (CM) itself is also an algorithm for the maximum mutual information classifications or estimations. Using the CM iteration algorithm [36], the Maximum Mutual Information (MMI) classification in the low-dimensional feature space is much comfortable. Combining the CM algorithm and the EM (or CM-EM algorithm) with the FCP, we can obtain a new mixture model-based classification method. It should be meaningful to apply the new convergence theory for mixture models and the CM iteration algorithm for MMI classifications to neural networks for machine learning.

## 7. Conclusions

The *Q-L* synchronization theory affirms that the complete data log-likelihood $Q$ and the observed data log-likelihood $L$ are positively correlated. Under the guidance of this theory, Ueda and Nakano proposed the DAEM algorithm, in which the annealing operation was used to avoid locally maximal $Q$. This paper used two examples, one of which was used by Ueda and Nakano, to show that the two affirmations in the *Q-L* synchronization theory are wrong. By analyzing the example used by Ueda and Nakano, this paper concluded that local convergence exists not because of locally maximal $Q$ but because of small samples and unfair competition between components. This paper hence proposed the FCP to reduce local convergence and accelerate global convergence.

This paper proposed a new convergence theory (the IEM theory) of mixture models. This theory claims that mixture models can converge not because $Q$ reaches its maximum but because information efficiency $G/R$ reaches its maximum, or $R$-$G$ reaches its minimum. The EM and CM-EM algorithm's convergence has been proved by the variational and iterative methods that Shannon *et al.* had used for analyzing the rate-distortion function. It has been proved that every step of the EM and CM-EM algorithms decreases the relative entropy $H(P||P_\theta)$, increasing the observed data log-likelihood.

About the EM algorithm's improvements, the main conclusions are:

- Using the FCP with the initialization map, we can select better initial means, especially for binary Gaussian mixtures.
- The CM-EM algorithm can accelerate the global convergence of the typical mixture models (in Examples 2-5) with locally maximal $Q$. On average, it can save 26% of the iterations and 18% of the running time that the EM algorithm needs.
- The new convergence theory reminds us to avoid blind improvements to the EM algorithm.

For more complicated mixture models, the author's study is not sufficient. However, we can expect that the IEM theory and the FCP could form a solid foundation for the EM algorithm's other improvements. It should be an excellent choice to learn from or combine some existing improved EM algorithms.

Two issues are attractive for further studies. One is to enrich the FCP and convert this principle into specific methods for more complicated mixture models' initializations. Another is to combine the new convergence theory and neural networks for machine learning.

**Appendix A.** Abbreviations

| Abbr. | Original text | Abbr. | Original text |
|---|---|---|---|
| EM | Expectation-Maximization | PMC | Population Monte Carlo |
| CM-EM | Channel matching EM | VEM | Variational EM |
| MM | Maximization-Maximization | SMEM | Splitting and merging EM |
| DAEM | Deterministic Annealing EM | CEM | Competitive EM |
| IEM | Information Efficiency Maximization | FCP | Fair Competition Principle |
| E3M | CM-EM whose E2-step repeats three times | MMI | Maximum Mutual Information |

**Appendix B. The Problem in Stanford University CS229-Part IX: The EM algorithm**

The proof can be simply stated as follows [29].

Let *x* be an observed data point, and *z* be a latent data point. According to Jensen's Inequality, we have

$$\begin{aligned}\log p(x;\theta) &= \log\sum_z P(x,z;\theta_i) = \log\sum_z Q(z)\frac{P(x,z;\theta_i)}{Q(z)} \\ &\geq \log\sum_z Q(z)\log\frac{P(x,z;\theta_i)}{Q(z)} = ELBO(x;Q,\theta),\end{aligned} \quad \text{(6,7,10 in [29])} \qquad \text{(a)}$$

where $Q(z)$ is *z*'s any function, "ELBO" means "evidence lower bound". Then we have

$$\begin{aligned}l(\theta) &= \sum_i \log p(x^{(i)};\theta) \geq \sum_x \log ELBO(x^{(i)};Q,\theta) \\ &= \sum_i Q_i(z^{(i)})\log\frac{p(x^{(i)},z^{(i)};\theta)}{Q_i(z^{(i)})}.\end{aligned} \quad \text{(11 in [29])} \qquad \text{(b)}$$

We simply set $Q_i(z^{(i)})=p(z^{(i)}|x^{(i)};\theta)$ so that $p(x,z;\theta)/Q(z)$ is constant (see Equation (8) in [29]). The parameters $\theta^{(t+1)}$ are then obtained by maximizing the right hand side of the equation above. Thus,

$$\begin{aligned}l(\theta^{(t+1)}) &\geq \sum_i \log ELBO(x^{(i)};Q^{(t)},\theta^{(t+1)}) \left(\text{because inequality } (11) \text{ holds for all } Q \text{ and } \theta\right) \\ &\geq \sum_i \log ELBO(x^{(i)};Q^{(t)},\theta^{(t)}) = l(\theta^{(t)}).\end{aligned} \qquad \text{(c)}$$

The above Inequality is between Equations (13) and (14) in [29].

QED.

The problem is that Inequality (c) is incorrect. We replace $l(\theta^{(i)})$ with $l(\theta^{(i)}|\theta^{(i)})$ *and* $l(\theta^{(i+1)})$ with $l(\theta^{(i+1)}|\theta^{(i+1)})$, and use $l(\theta^{(i+1)}|\theta^{(i)})$ to denote *l*(new *θ for P(x|θ)*| last *θ for P(y|x)*). Accorording to Inequality (11) in [29], we can only obtain

$$l(\theta^{(t+1)}) = l(\theta^{(t+1)}|\theta^{(t+1)}) \geq \sum_i \log ELBO(x^{(i)};Q^{(t+1)},\theta^{(t+1)}); \qquad \text{(d)}$$

$$l(\theta^{(t+1)}|\theta^{(t)}) \geq \sum_i \log ELBO(x^{(i)};Q^{(t)},\theta^{(t+1)}). \qquad \text{(e)}$$

There is no reason in [29] for

$$\sum_i \log ELBO(x^{(i)};Q^{(t+1)},\theta^{(t+1)})\text{(after the E-step)}$$
$$\geq \sum_i \log ELBO(x^{(i)};Q^{(t)},\theta^{(t+1)})\text{(before the E-step)} \tag{f}$$

so that $l(\theta^{(i+1)}|\ \theta^{(i+1)})\geq l(\theta^{(i)}|\theta^{(i)})$. Inequality (f) is similar to Inequality (13) in the main text. We need to prove that the E-step makes them tenable. However, there is no such proof in [29]. Without such proof, we cannot understand why the E-step uses $P(y|x)$ like that or why we use $Q_i(z^{(i)})=p(z^{(i)}|x^{(i)};\theta)$.

**Appendix C. The Proof of $\frac{dH(Y|X,\theta)}{d\sigma}>0$ for Example 1.**

Letting $t=P(y_1|x)$, we have

$$\frac{dH(Y|X,\theta)}{dt}=\sum_i P(x_i)\frac{d}{dt}[-t\log t-(1-t)\log(1-t)]$$
$$=\sum_i P(x_i)\log\frac{1-t}{t}=\sum_i P(x_i)\log(e^{-b/\sigma^2})=-\sum_i P(x_i)b/\sigma^2.$$

Since

$$\frac{dt}{d\sigma}=\frac{d}{dt}(\frac{1}{1+e^{-b/\sigma^2}})=-\frac{2be^{-b/\sigma^2}\sigma^{-3}}{(1+e^{-b/\sigma^2})^2},$$

we have

$$\frac{dH(Y|X,\theta)}{d\sigma}=\frac{dH(Y|X,\theta)}{dt}\frac{dt}{d\sigma}=\sum_i P(x_i)\frac{2b^2e^{-b/\sigma^2}\sigma^{-5}}{(1+e^{-b/\sigma^2})^2}>0.$$

**Appendix D: Optimizing $P(Y|X)$ and $P(Y)$ to minimize $R$-$G$**

To optimize $P(Y|X)$, order

$$\frac{\partial F}{\partial P(y_j|x_i)}=\frac{\partial}{\partial P(y_j|x_i)}\{\sum_j\sum_i P(x_i)P(y_j|x_i)\log\frac{P(y_j|x_i)}{P(y_j)}$$
$$-\sum_j\sum_i P(x_i)P(y_j|x_i)\log\frac{P(x_i|y_{\theta j})}{P(x_i)}+\mu_i\sum_j P(y_j|x_i)+\alpha\sum_j P(y_j)\}=0.$$

Hence

$$P(x_i)[1+\log P(y_j|x_i)]-P(x_i)\log[P(x_i|y_{\theta j})/P(x_i)]+\mu_i=0,$$
$$\log[P(y_j|x_i)/P(y_j)]=\log[P(x_i|y_{\theta j})/P(x_i)]-(\mu_i+1)/P(x_i).$$

Order $\log\lambda_i=(\mu_i+1)/P(x_i)$, we have

$$P(y_j|x_i)=P(y_j)P(x_i|y_{\theta j})/\lambda_i,\ i=1, 2, \ldots, n;\quad j=1, 2.$$

Since the second-order partial derivative is greater than 0, this $P(Y|X)$ minimizes $I(X; Y)$-$I(X; \theta)$. Since $P(y|x_i)$ is normalized, the optimized $P(Y|X)$ is

$$P^*(y_j|x_i)=P(y_j)P(x_i|y_{\theta j})/\sum_k P(y_k)P(x_i|y_{\theta k}),\ i=1, 2, \ldots, n;\quad j=1, 2.$$

To optimize $P(Y)$, order

$$\frac{\partial F}{\partial P(y_j)} = \frac{\partial}{\partial P(y_j)}[\sum_j \sum_i P(x_i)P(y_j \mid x_i)\log\frac{P(y_j \mid x_i)}{P(y_j)}$$
$$+\sum_i \mu_i \sum_j P(y_j \mid x_i) + \alpha \sum_j P(y_j)] = 0$$
.

Hence

$$-\sum_i P(x_i)P(y_j \mid x_i)/P(y_j) + \alpha = 0,$$

$$P(y_j) = \frac{1}{\alpha}\sum_i P(x_j)P(y_j \mid x_i).$$

Since the second-order partial derivative is greater than 0, this $P(Y)$ minimizes $I(X;Y)$-$I(X;\theta)$. Since $\sum_j P^*(y_j)$=1, $\alpha$=1. Therefore, the optimized $P(y)$ is

$$P^*(y_j) = \sum_i P(x_i)P(y_j \mid x_i),\ j=1, 2, \ldots, n.$$

## Supplemental Materials

The supplemental materials, including several Python 3.6 source codes for Figures 1, 2, 5, 7-12, can be downloaded from http://survivor99.com/lcg/cm/PythonMM4Ex1-6.zip.

**Table 1.** The list of Supplemental Materials

| File folders | Python Files |
|---|---|
| Ex1-greaterQ | For Figure 1 and Figure 2 |
| Ex2-localCon | For Figure 5 and Figure 7 |
| Ex3-g2-1-50 | For Figure 8 |
| Ex3-g2-2-50 | For Figure 9 |
| Ex5-2D-fair | For Figure 10 |
| Ex6-2D-s3-15 | For Figure 11 and Figure 12 |
| Compare-speeds | For compariing speeds in Section 5.3 |

**Author Contributions:** Chenguang Lu is the only author for all work, including programming.

**Funding:** This research received no external funding.
**Acknowledgments:** The author thanks …
**Conflicts of Interest:** The author declares no conflict of interest.